# Automated Image-Based Identification and Consistent Classification of Fire Patterns with Quantitative Shape Analysis and Spatial Location Identification

Pengkun Liu [a], Shuna Ni [b], Stoliarov, Stanislav I [b]. and Pingbo Tang [a]*

[a] *Department of Civil and Environmental Engineering, Carnegie Mellon University, 5000 Forbes Avenue, Pittsburgh, PA, 15213, United States*
[b] *Department of Fire Protection Engineering, University of Maryland at College Park, 4356 Stadium Drive, College Park, MD 20742, United States*

* Corresponding Author: ptang@andrew.cmu.edu

## Abstract

Fire patterns, consisting of fire effects that offer insights into fire behavior and origin, are traditionally classified based on investigators' visual observations, leading to subjective interpretations. This study proposes a framework for quantitative fire pattern classification to support fire investigators, aiming for consistency and accuracy. The framework integrates four components. First, it leverages human-computer interaction to extract fire patterns from surfaces, combining investigator expertise with computational analysis. Second, it employs an aspect ratio-based random forest model to classify fire pattern shapes. Third, fire scene point cloud segmentation enables precise identification of fire-affected areas and the mapping of 2D fire patterns to 3D scenes. Lastly, spatial relationships between fire patterns and indoor elements support an interpretation of the fire scene. These components provide a method for fire pattern analysis that synthesizes qualitative and quantitative data. The framework's classification results achieve 93% precision on synthetic data and 83% on real fire patterns.

## 1. Introduction

A fire pattern also called a burn pattern, combines fire effects and associated characteristics that produce a measurable and repeatable form [1]. Fire effects are the observable or measurable changes in or on a material or system due to fire [1,2]. Accurate interpretation of fire patterns can aid in determining the fire's origin. Since the beginning of organized fire investigation in the late 1940s, fire investigators have relied on fire patterns as their basis to determine the origin of the fire [3]. Fire patterns can be classified as plume-generated, ventilation-generated, hot gas layer-generated, full room involvement-generated, and suppression-generated [3]. According to the geometric shape, a plume-generated fire pattern can be classified into different types, such as inverted cone patterns, V patterns, U-shaped patterns, hourglass patterns, pointer and arrow patterns, and circular-shaped patterns.

The shape of a fire pattern also provides information related to fire dynamics. In the current practices of fire investigation, the interpretation of fire patterns is implicit and subject to investigator bias, heavily relying on fire investigators' knowledge, experience, education, training, and skill without the benefit of a structured framework to help guide fire investigators through the process. These biases and subjectivity can lead to a misinterpretation of fire patterns, casting doubt on the validity of the conclusions of fire investigation and on the ability of the process to meet the Daubert standards [4,5]. Consequently, it is necessary to develop well-validated methods to analyze fire pattern data quantitatively, consistently, and automatically.

Some research has focused on solving the problem mentioned above. For example, Gorbett et al. [6] proposed a methodology that involved dividing a photograph of a fire pattern into small gridded sections and then, by visual inspection, classifying the damage degree in each cell into one of seven damage classes. However, despite using numerical indices, the classification of fire damage degrees is still qualitative, relying on visual inspection, and is therefore subjective. There is a notable lack of robust methodologies that systematically classify a fire pattern's shape. Additionally, existing studies do not adequately cover the representation of spatial relationships among fire patterns and their connections with nearby objects or furniture within a fire scene. This limitation is significant, as the spatial distribution of evidence in a fire scene can provide crucial clues about the fire's origin and progression.

Therefore, this paper aims to qualitatively advance the consistent classification of plume-generated fire patterns. The study tackles crucial research questions:
1. What representations capture the qualitative semantics of fire patterns to help consistent fire pattern classification?
2. What image analytics can accurately identify correlated spatial fire patterns for precise location identification?

Addressing these queries introduces several technical hurdles. First, due to the intricate nature of fire patterns, to the best of the authors' knowledge, no quantitative and consistent representation of fire patterns exists to aid their classification and improve the comprehension of a fire scene. Second, an image analytics method is needed to examine correlated spatial fire patterns and enhance the reliability of fire investigations. Although 2D fire pattern images can effectively capture the colors and textures of fire-damaged surfaces, they fail to provide sufficient information about the absolute geometric characteristics of a fire pattern and its correlations with other evidence and features of a fire scene, which are crucial for a comprehensive understanding of a fire pattern. For example, defining whether a fire pattern is ventilation-induced requires understanding its spatial relationship with an opening. This limitation can be addressed with 3D imaging techniques such as laser-scanning point clouds that provide precise geometries of captured objects. Furthermore, a relationship analysis of various fire patterns from point clouds is essential for achieving practical, explainable pattern analysis, aiding fire investigators in studying fire scenes.

This study proposes an automated framework for identifying and classifying fire patterns through image analysis. This framework employs quantitative shape assessments and spatial relationships analysis, offering an approach to understanding fire scenes. At its core, the framework integrates several vital components: First, it leverages human-computer interaction (HCI) mechanisms, enabling users to recognize and accurately extract fire patterns from 2D images. This interaction is critical to ensure the system can adapt to various fire scenarios and the complexities of patterns. Secondly, the framework introduces quantitative definitions and classifications for fire patterns. By quantifying these patterns, the system provides a more objective and scalable means of categorization, which is essential for consistent analysis of different fire incidents. The third component focuses on segmenting point clouds obtained from indoor fire scenes. This process is vital for creating detailed three-dimensional representations of a fire scene, facilitating a deeper understanding of fire development in a scene. Lastly, the framework enhances the analysis of a fire scene by projecting classified fire patterns back to the 3D representation of the fire scene where those patterns exist and integrating this information with a scene graph that represents the spatial relationship between fire patterns and other evidence and features of the scene.

This research sets specific objectives and contributions: (1) it seeks to define the quantitative representation of fire patterns using aspect ratios. This approach involves comparing the aspect ratios of different fire patterns to facilitate their classification on a quantitative basis and enhance the accuracy and consistency of fire pattern classifications. (2) the study aims to develop methodologies for extracting and analyzing fire patterns' spatial location identification using 2D and 3D images. Doing so is expected to achieve a more nuanced understanding of fire incidents, linking quantitative data and visual insight with contextual information. This dual-pronged approach is expected to enhance the quantitative interpretation of fire patterns and scenes.

The subsequent sections are organized as follows: Section 2 reviews the literature. Section 3 delineates the proposed automated and quantitative image-based fire pattern analysis methodology. Section 4 describes the fire pattern recognition and extraction methodology through human-computer interaction. Section 5 presents a quantitative definition and classification of fire patterns with physical features. Section 6 introduces point-cloud segmentation for analyzing indoor fire scenes. Section 7 presents the understanding of the fire scene by generating and analyzing scene graphs. Section 8 concludes with comments and directions for future research.

## 2. Literature Review

### 2.1 Fire Pattern Definition and Challenges

Fire patterns are one type of evidence that fire investigators use to trace the origin and progress of a fire. Fire patterns can be classified by their generation or causal relationship to fire dynamics by providing the following classes: plume-generated patterns, ventilation-generated patterns, hot gas layer-generated patterns, full-room involvement-generated patterns, suppression-generated patterns, and undetermined-generated patterns [1]. Plume-generated patterns are the most common type of fire patterns. Being three-dimensional, fire plumes create patterns that intersect with two-dimensional surfaces like ceilings or walls. Plume-generated patterns [7] can be described as the following shapes: inverted cone patterns (Figures 1a and 1e), V patterns (Figures 1b and 1f), U-shaped patterns (Figures 1c and 1g), hourglass patterns, pointer and arrow patterns, and circular-shaped patterns. Additionally, rectangular (columnar) fire patterns can also be observed in some scenarios. Here, the inverted cone pattern is characterized by an inverted triangle shape on vertical surfaces near the fire's origin, and this pattern forms in the fire's early stages [7]. The pattern's apex points upward, reflecting the natural spread of the flame upward. The inverted-cone pattern indicates the fire's point of origin, with the triangle's base suggesting the area where the fire started. As the heat release rate and flame height increase, this pattern can evolve into a rectangular shape (Figs. 1d and 1h). The rectangular pattern is a vertical pattern with a uniform width that forms as the fire intensifies, creating a direct upward thermal current before significant outward spread. It suggests a localized, intense fire, pointing towards a stage of development where the fire interacts with ceilings or overhanging structures. A V-shaped pattern becomes defined as the fire spreads upward and outward. The investigator is cautioned that a V shape is not necessarily associated with the fire's origin as it can also be created due to protected areas from contents, combustion due to ventilation flow paths, and dropdown. The U-shaped patterns form when a fire plume located some distance from a wall interests the wall.

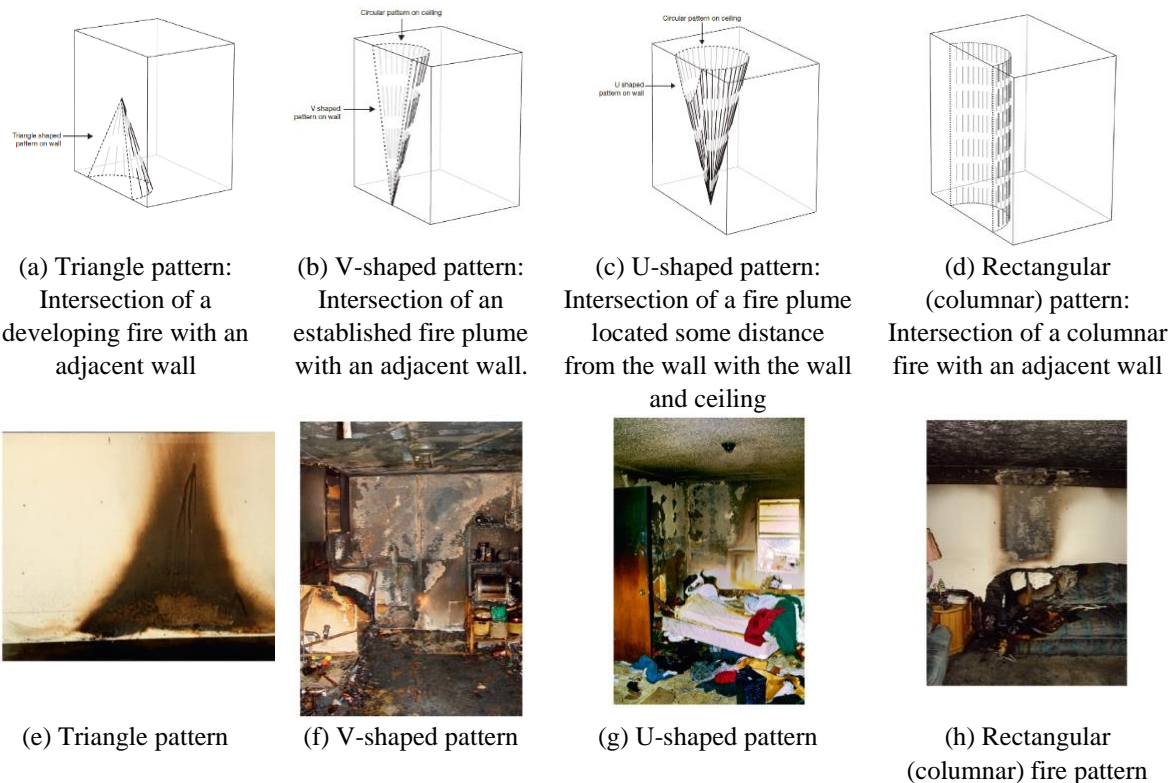

(a) Triangle pattern: Intersection of a developing fire with an adjacent wall

(b) V-shaped pattern: Intersection of an established fire plume with an adjacent wall.

(c) U-shaped pattern: Intersection of a fire plume located some distance from the wall with the wall and ceiling

(d) Rectangular (columnar) pattern: Intersection of a columnar fire with an adjacent wall

(e) Triangle pattern

(f) V-shaped pattern

(g) U-shaped pattern

(h) Rectangular (columnar) fire pattern

Figure 1. Classification of plume-generated fire patterns [7]

The misinterpretations and misclassifications of fire patterns might lead to inaccurate determinations of fire origins, which is critical in arson investigations. Despite the essential nature of this work, not every fire investigator has equal levels of education, training, or understanding of how fires interact with their surroundings. Much of their expertise appears to be derived from informal on-the-job training, where more experienced investigators train their less experienced counterparts, sometimes perpetuating outdated or incorrect techniques and theories [3]. The problem of insufficient knowledge transfer is primarily attributed to the reliance on outdated information by experienced investigators [8]. A tragic manifestation of this knowledge gap was the execution of Cameron Todd Willingham by the State of Texas, based on an investigation that was critically flawed due to the outdated understanding of fire science and the failure to acknowledge the limitations of modern fire investigation [9].

## 2.2 Fire Patterns Classification with Spatial Relationship

Fire-pattern classification analyzes a fire pattern regarding its shape and spatial relationship to growing fire. These classification results can help infer the fire dynamics underlying fire patterns. Fire patterns need to be analyzed based on both 2D and 3D images. The former type of image effectively captures the colors and textures of fire-damaged surfaces but cannot provide sufficient information to identify the absolute geometric characteristics of a fire pattern and its causal relationship to fire dynamics. Many fire patterns are spatially distributed, and their spatial relationship to fire dynamics usually cannot be identified purely through their own attributes. For example, the spatial relationships of an opening with a fire pattern are necessary to identify whether a fire pattern is ventilation-generated. This limitation can be overcome by 3D images (e.g., 3D laser-scanning point clouds) providing the absolute geometries of captured objects. Accordingly, it is significant to develop fire-pattern classification models using integrated analysis of 3D and 2D images.

As for 2D fire pattern analysis, methodologies or techniques that make fire investigation more objective and quantitative can help somewhat alleviate the aforementioned problems. For example, fire pattern analysis can be more quantitative, which is a significant gap in investigation methodologies [10]. However, advances in computer science, particularly in shape analysis and pattern recognition, offer promising avenues for addressing this challenge [11–13]. Several studies have tried to detect and classify fire scenes [11–13] with technologies such as deep learning-based image or video-based fire classification in the domain of fire detection for indoor or outdoor environments and so on [14–22]. However, they have not been used in fire pattern classification and analysis. Furthermore, most of these methods lack the explanatory and physical description of fire patterns. Fire investigation is a legally sensitive domain; it needs a reason to classify fire patterns. Methods to quantitatively describe shapes in computer science, such as geometric morphometry [23], contour analysis [24], and machine learning algorithms [25], have the potential to provide objective, measurable data rather than relying solely on subjective judgment. Shape representation generally looks for practical and perceptually important shape features based on shape boundary information or boundary plus interior content. Various features have been designed, including shape signature, signature histogram, shape invariants, moments, curvature, shape context, shape matrix, spectral features, etc. These various shape features are often evaluated by how accurately they allow one to retrieve similar shapes from a designated database [24]. These existing techniques have the potential to be applied in fire pattern classification.

Recent technological advancements have significantly improved traditional fire scene analysis methods. The development and application of 3D mapping technologies, utilizing drones equipped with LiDAR (Light Detection and Ranging) or photogrammetry, have revolutionized this field [26–30]. These 3D models provide a comprehensive view of the pre-fire scene, active fire spread, and post-fire damage, allowing better monitoring or reconstruction of wildfire spread and better damage assessment [26–30]. This technology aids real-time planning and execution of firefighting strategies and supports post-fire analysis [11,29–32]. Based on the significant role of 3D images in outdoor wildfire analysis, 3D imaging is also expected to be highly beneficial for indoor fire scene analysis. For example, they provide realistic absolute geometric information about fire patterns and enable the analysis of spatial relationships among fire patterns and between fire patterns and other objects. This facilitates a better understanding of how fire patterns relate to the burning of other objects and how they are influenced by objects in the scene, thereby enhancing the understanding of fire development. However, such research is currently lacking for indoor fire scene analysis.

## 2.3 Research Gaps and Motivation Case for this study

This review has identified the following critical research gaps: First, current research does not adequately address the quantitative representation of fire patterns. There is a notable lack of robust methodologies that systematically classify fire patterns based on geometric data. This gap impedes the ability to standardize fire investigations. Furthermore, existing studies do not adequately cover the representation of spatial relationships between fire patterns and their interaction with nearby objects (e.g., furniture) within a fire scene. This limitation is a significant oversight, as the spatial arrangement of a fire scene can often provide critical clues regarding the origin and progression of a fire. These gaps undermine the accuracy of fire investigations and increase the likelihood of errors, as discussed in Section 1. Enhancing fire pattern analysis with better quantitative tools and methods to assess spatial relationships will lead to more reliable and effective fire investigations.

## 3. Methodology - Automated and Quantitative Image-Based Fire-Pattern Analysis

This framework consists of four steps, as shown in Fig. 2: (1) human-computer interaction for fire pattern recognition and extraction; (2) fire patterns quantitative definition and classifications; (3) indoor fire scenes point clouds segmentation; (4) fire scene understanding through the projection of the classified fire patterns onto a 3d fire scene and scene graph.

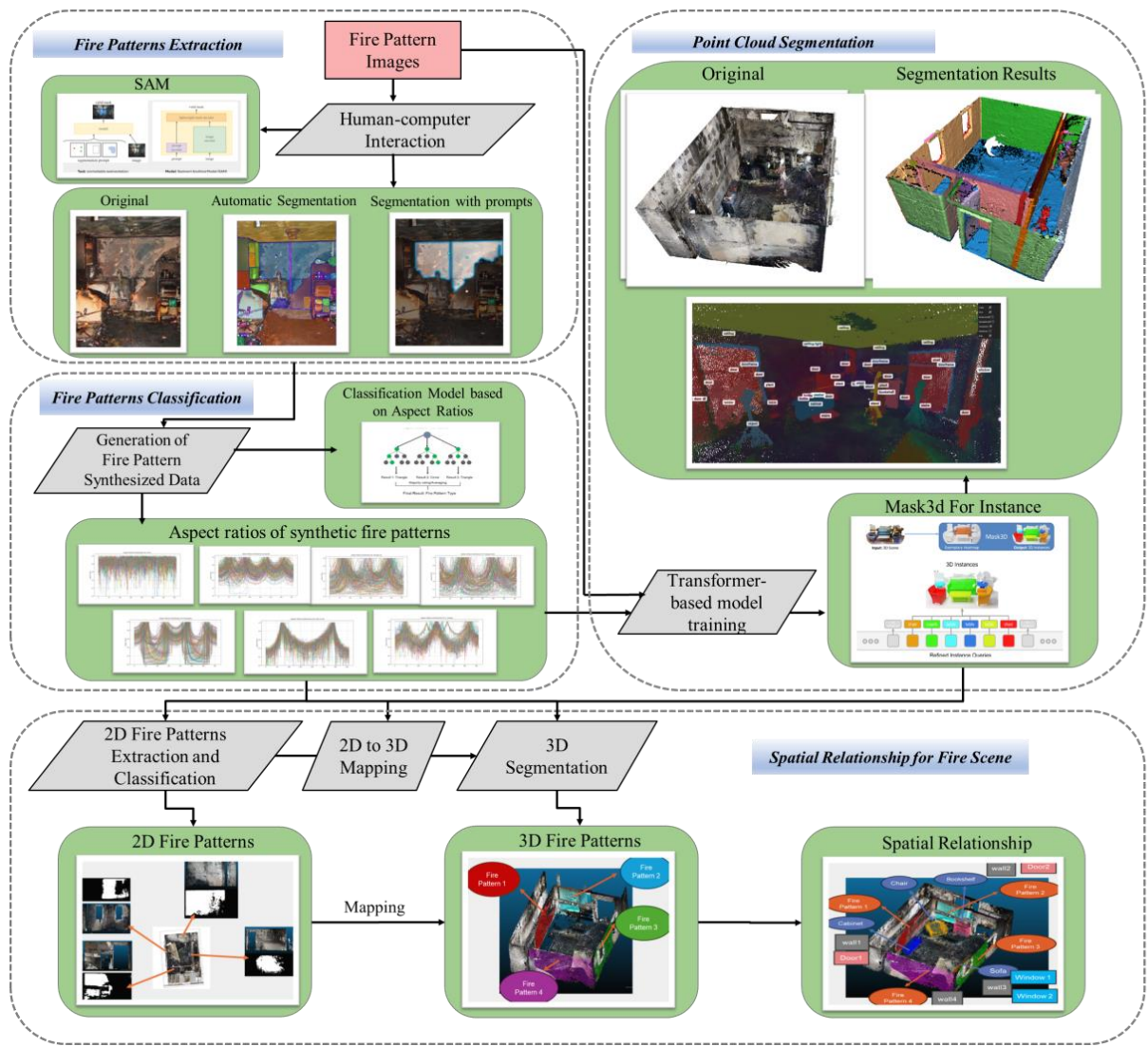

Figure 2. Framework for Automated and Quantitative Image-Based Fire-Pattern Analysis

**Human-Computer Interaction for Fire Pattern Recognition and Extraction**: This step involves integrating fire investigators and computers for accurate fire pattern recognition and extraction. Through human-computer interaction, investigators input their observations and expertise into the system, using algorithms to recognize and extract fire patterns from the data. This process could involve using graphical interfaces to annotate and highlight areas of interest in fire scene images. These annotations then serve as training data for AI, enabling it to learn from expert inputs and gradually automate the segmentation of fire patterns with greater accuracy.

**Fire Patterns Quantitative Definition and Classification**: This phase focuses on defining and classifying fire patterns based on their geometric shapes, such as triangular, rectangular (columnar), or conical. The shape of a fire pattern, as discussed in Section 2.1, reveals information about fire dynamics and development, thus supporting the data analysis of fire investigation.

**Indoor Fire Scenes Point Clouds Segmentation**: Point-cloud segmentation involves processing 3D scans of fire scenes to identify and segment different structures and objects within the scene.

A point cloud is a set of data points in space produced by 3D scanners that measure many points on the external surfaces of objects around them.

**Fire Scene Understanding Through Spatial Relationship**: Scene graphs represent a method of structuring and interrelating data to improve the understanding of complex information. In the context of fire scene analysis, a scene graph can help map the relationships and dependencies between fire patterns and environmental factors. For example, it could help determine whether a particular fire pattern was caused by ventilation (ventilation-generated) or by a plume (plume-generated). By analyzing the spatial interplay between fire patterns and factors such as compartment openings or furniture placement, investigators can gain deeper insights into how a fire starts and progresses.

By leveraging human-computer interaction, quantitative classification, point cloud segmentation, and scene graphs, the framework aims to enhance fire investigations' accuracy, objectivity, and comprehensiveness. Our approach extracted fire patterns from 2D images, and 3D laser scanning point clouds classified them. The synthesized data, along with human-computer interaction, improved the accuracy and reliability of the extraction process. Projecting 2D fire patterns into the 3D point-cloud scene provided a more holistic view of the fire-related damage. Geolocating fire damage within the building's coordinate system improves the understanding of fire progression and damage assessment.

## 4. Fire Pattern Extraction Through Human-Computer Interaction

## 4.1 Framework for Fire Pattern Extraction through Human-Computer Interaction

The framework of fire pattern extraction through HCI is the Segment Anything Model (SAM). [33], which is a framework designed for applications in human-computer interaction, particularly for the recognition and extraction of fire patterns, as shown in Fig. 3. This framework is structured around two primary components: This first one is the prompt segmentation task. This component is essential for achieving zero-shot generalization. It allows this framework to perform segmentation tasks effectively across various fire contexts without requiring task-specific training. The other is Model Architecture. The architecture supports real-time performance and scalability while utilizing powerful pre-training techniques that are standard in transformer-based vision models. SAM is distinguished by its training regimen, which includes millions of images and more than a billion masks, preparing the model to respond accurately to various segmentation prompts. These prompts could be as simple as points indicating foreground and background or as complex as textual descriptions, all serving as inputs that guide the segmentation process. This versatility is crucial for tasks requiring precise and adaptive segmentation capabilities, such as detecting and analyzing fire patterns.

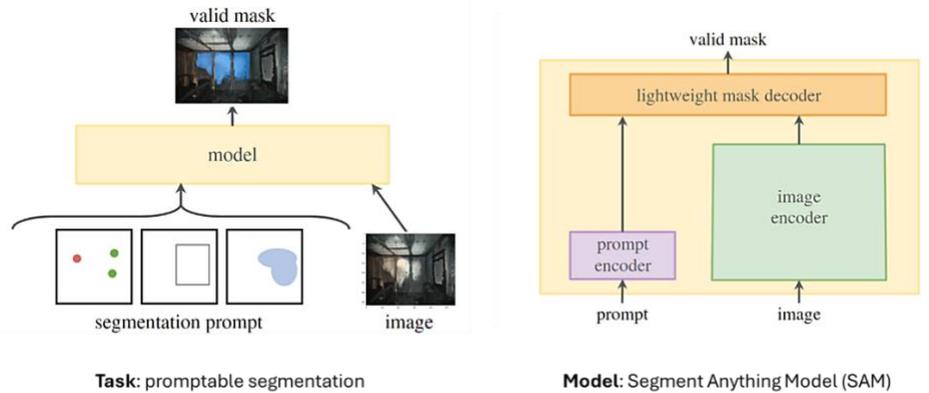

Figure 3. Framework of Fire Pattern Extraction through Human-Computer Interaction [33]

The operation of the SAM framework is delineated into several distinct stages: (1) **Input stage:** In this initial stage, the user provides an image or a set of images containing fire patterns that they wish to recognize or extract. The user might also provide input prompts specifying the particular areas of interest in the fire patterns. (2) **Segmentation stage:** This stage utilizes the powerful capabilities of SAM. The framework processes the input image(s) and identifies and segments the fire patterns based on the user's prompts. SAM produces high-quality object masks from these prompts, which could be as simple as points or boxes. This segmentation is achieved by recognizing the features that define fire patterns and creating boundaries or segments around these patterns, effectively isolating them from the rest of the image. (3) **Extraction and Recognition Stage**: After segmenting the fire patterns, they are extracted from the original image for further analysis or use. (4) **Interaction Stage**: The results of the segmentation, extraction, and recognition processes are then presented to the user. Here, the human-computer interaction aspect of the framework shines as the user can interact with these results to adjust parameters, refine the segmentation, provide feedback to the model, or use the results for further analysis. (5) **Feedback loop:** User feedback, a critical component of HCI, is used to improve the model performance. This valuable feedback is integrated into the model's training, refining its ability to segment, recognize, and extract fire patterns in future tasks. This creates a feedback loop where the user's input continuously improves the model's performance.

## 4.2 Segment Anything Framework Architecture

The SAM network detailed architecture contains three crucial components, the image encoder, the prompt encoder, and the mask decoder. The input is a fire pattern image; the prompt could be fire investigator instructions, and the outputs are valid masks representing segmented fire patterns with confidence scores. At its core, the Image Encoder utilizes a Masked Auto-Encoder (MAE), a pre-trained Vision Transformer (ViT), to produce unique image embeddings in a single pass. This encoder efficiently handles high-resolution input and is operational before the model receives prompts, facilitating smooth integration into the segmentation workflow.

The prompt encoder converts inputs such as points, masks, bounding boxes, or textual prompts into corresponding embedding vectors in real time. The architecture distinguishes between two types of prompts: Sparse prompts (points, boxes, text), which utilize positional encodings enhanced with learned embeddings specific to each prompt type. Dense prompts

(masks) are processed through convolutions and are combined elementwise with the image embeddings. The Prompt Encoder effectively manages these diverse inputs, ensuring that whether the prompts are dense or sparse, they integrate seamlessly with the image embeddings to maintain the model's adaptability and response efficiency.

The Mask Decoder is a streamlined component designed to predict segmentation masks from the combined embeddings produced by the Image and Prompt Encoders. A modified decoder block inspired by existing Transformer decoder architectures incorporates prompt self-attention and cross-attention mechanisms. These attention processes facilitate a dynamic and precise mapping of embeddings to segmentation masks, enhancing the model's real-time performance and interactive capabilities.

## 4.3 Fire Pattern Extraction Results Through Human-Computer Interaction

Fig. 4 shows the segmentation capabilities of the SAM when applied to diverse fire patterns observed in indoor settings. The model demonstrates proficiency in autonomously dissecting distinct components and isolating fire patterns from the imagery, as depicted in Fig. 6(a) and (b). However, this automatic segmentation encounters limitations when faced with complex backgrounds, where it fails to accurately disengage the fire patterns on the walls. In such complex scenarios, augmenting SAM with human intervention proves beneficial. The model, which initially might only recognize the wall as a monolithic object, disregarding the nuanced patterns formed by fire, such as those with rectangular or inverted-triangle shapes shown in Figs. 4 (c) and (d), evolves with human input. Adding human prompts, indicated by strategic points representing targeted instructions, enables the SAM to discern and extract intricate fire patterns amidst the noisy and chaotic environments typical of indoor fire scenes. This synergy of SAM and human insight significantly enhances the model's efficacy, allowing for a more nuanced analysis and interpretation of fire-affected environments.

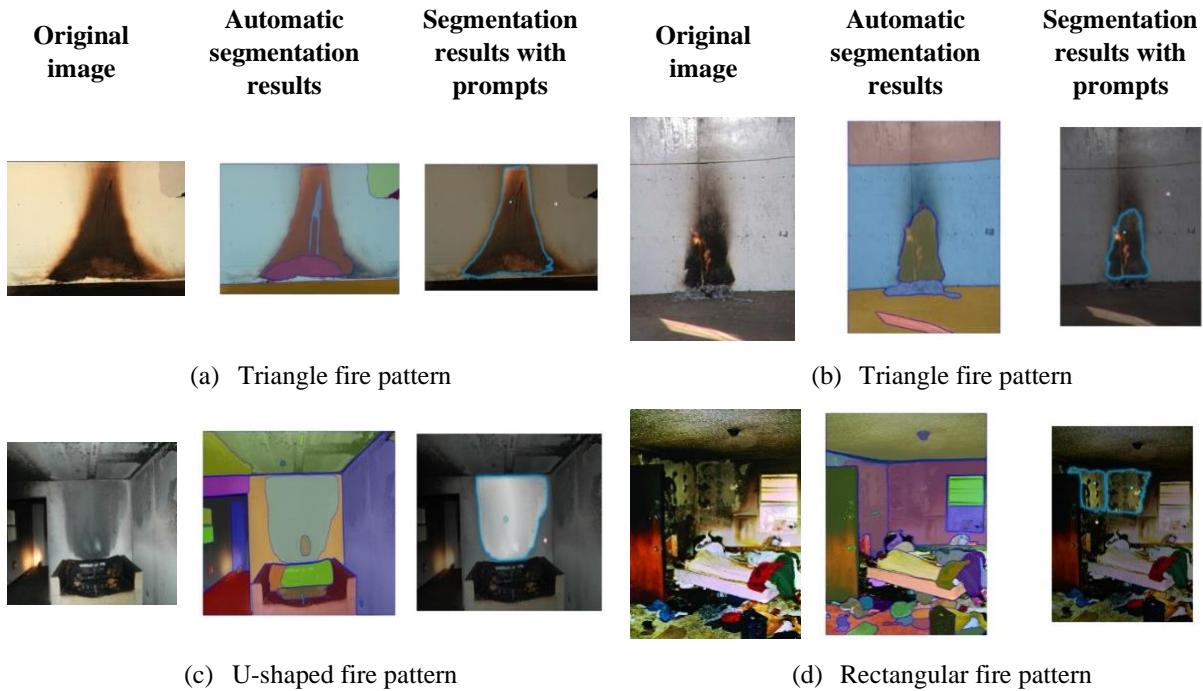

(a) Triangle fire pattern      (b) Triangle fire pattern

(c) U-shaped fire pattern      (d) Rectangular fire pattern

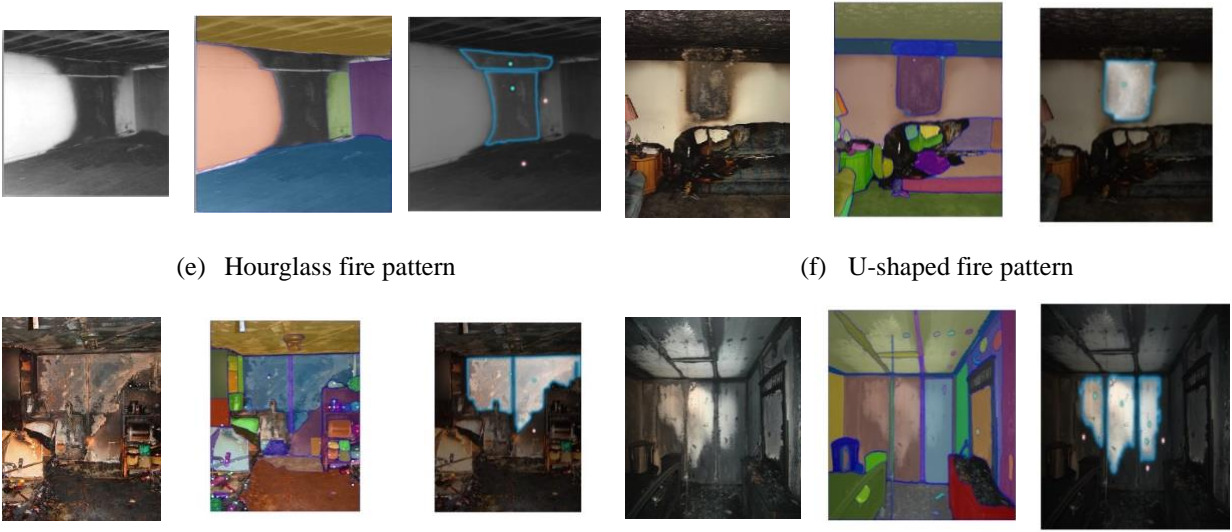

(e) Hourglass fire pattern   (f) U-shaped fire pattern

(g) V-shaped fire pattern   (h) V-shaped fire pattern

Figure 4. Fire Pattern Extraction Results

## 5. Quantitative Definition and Classification of Fire Patterns

### 5.1 Fire Pattern Quantitative Definition of Fire Patterns by Aspect Ratios

In Fig. 5, we introduce a methodology for quantifying fire patterns based on their geometric characteristics by defining their aspect ratios. This quantitative analysis allows for a standardized comparison and a detailed description of a fire pattern. The following steps outline the procedure for calculating the aspect ratios: (1) centroid determination: The preliminary step in the analysis involves locating the centroid of each extracted fire pattern. This is achieved by calculating the meaning of the coordinates for all sample points within the shape. (2) Construction of a rotation line: A vertical line is constructed through the shape's centroid. This line intersects the shape at two points, and the length of the line segment within the shape envelope is recorded for further analysis. (3) Line rotation procedure: With the centroid as the fulcrum, the line is rotated incrementally counterclockwise around the centroid of a shape within a two-dimensional plane. The length of a line segment intersecting the shape is carefully recorded for each rotation angle. (4) Calculation of the aspect ratio: The aspect ratio is calculated as the length of the line segment obtained in Step 3 divided by the maximum length of any segment observed during the entire rotation. (5) Aspect ratio recording: The aspect ratios for each rotational position are stored systematically in a list. This collection becomes a comparative tool to analyze fire patterns. A consistent aspect ratio at various angles may signify a pattern with stable geometric properties, whereas fluctuating ratios may indicate a pattern with varying geometric features.

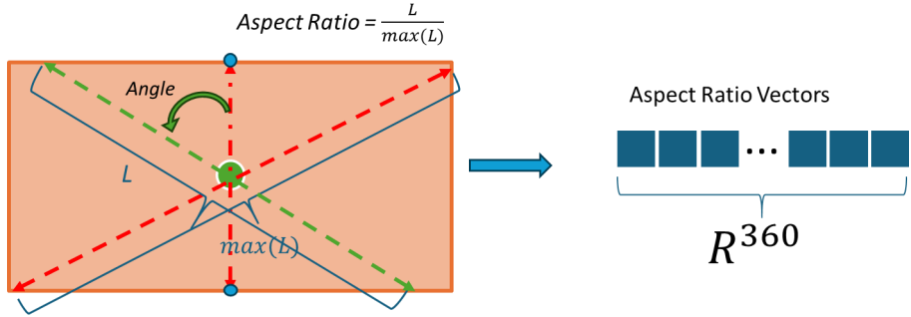

Figure 5. Illustrations of aspect ratio for shape descriptions

Fig. 6 shows the aspect ratios generated for some standard fire patterns as functions of the rotational angles (starting from the horizontal line measured in degrees from 0 to 360). In this paper, we do not consider the pointer and arrow patterns. In NFPA 921, the circular-shaped pattern includes both complete circular and partial patterns. In our methods, however, we distinguish between the two due to the significant differences in their aspect ratios. The graphs show significant variations for each shape regarding the peaks and valleys' numbers, positions, and numerical ranges. Peaks in the aspect ratio are points on the edge of a fire pattern that represents the maximum outward extension within a specific segment. Valleys in the aspect ratio are points that denote the maximum inward recession within a segment of a fire pattern. For example, the circle maintains a constant aspect ratio, as indicated by a flat line on the graph, which means that the aspect ratio remains unchanged regardless of the angle of the line segment. The rectangle shows four distinct peaks and three valleys in its aspect ratio graph, corresponding to the diagonals of a rectangular, and the line segments parallel to the short side. The aspect ratio graph of the U shape displays a wavelike pattern, with two distinct peaks and three valleys corresponding to the changing lengths of the rotating line segment as it intersects the varying widths of the shape. The triangle-down shapes present a pattern of two peaks and three valleys.

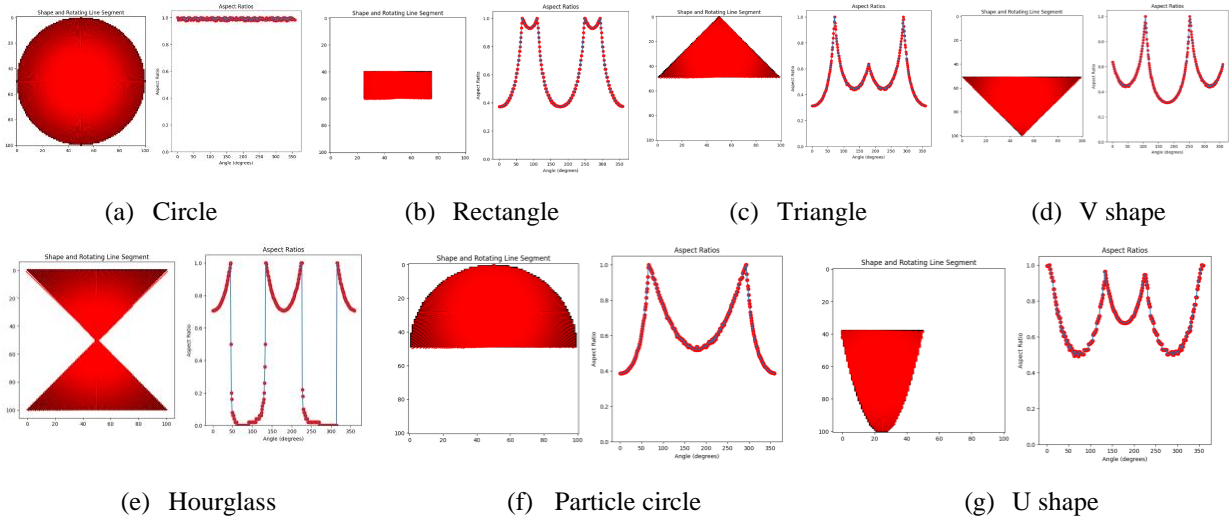

(a) Circle     (b) Rectangle     (c) Triangle     (d) V shape

(e) Hourglass     (f) Particle circle     (g) U shape

Figure. 6. Illustrations of aspect ratio for defined standard fire patterns

Triangle-up shapes display three high points and two low points, demonstrating variations in the aspect ratio as the line follows the base and then the sides of the triangle first. Hourglass figures produce a consistent four-high-point and three-low-point rhythmic pattern due to oscillations in the aspect ratio caused by the alternating wide and narrow segments of the hourglass form, manifesting the shape's symmetry. The aspect ratio graph for partial circles outlines an M-like fluctuation. When assessing realistic fire patterns from images, we can categorize their shapes and link them to certain standard fire configurations by inspecting the distributed aspect ratio and contrasting it with these established forms. For a realistic fire pattern extracted from images, we can classify its shape and associate it with a specific standard fire pattern by analyzing its aspect ratio distribution and comparing it to these standard shapes. This method offers a quantitative dimension to qualitative analysis, potentially enhancing precision and uniformity in evaluating fire patterns during investigations.

## 5.2 Generation of Synthetic Data for Supporting Fire Pattern Analysis.

### 5.2.1 Generation of Synthetic Data

Due to the limited fire pattern images from natural indoor fire scenes, we generated the fire pattern synthesized dataset to increase the diversity of the fire pattern dataset and prepare for a robust fire pattern classification. Fig. 7 shows a collection of shapes designed to represent various fire patterns, improved with edge smoothing, noise addition, and distortion to simulate the imperfections characteristic of actual fire damage. This process is instrumental in creating a comprehensive dataset that trains algorithms to recognize and classify real-world fire patterns. As shown in Fig. 7., the image arranges the shapes in rows, each depicting a particular category of fire patterns: circles, rectangular, triangles (oriented upward and downwards), half-circles, U shapes, and hourglasses. For each category, Fig. 7 (a) presents the original shapes, and Fig. 7 (b) presents the shapes with edge smoothing and noise. Edge smoothing is likely to replicate the effects of heat and combustion on edges and surfaces. At the same time, introducing noise replicates the randomness and irregularity that fire imparts to objects and structures. This extensive array prepares a machine-learning model for the diverse and unpredictable fire damage encountered in actual fire scenes.

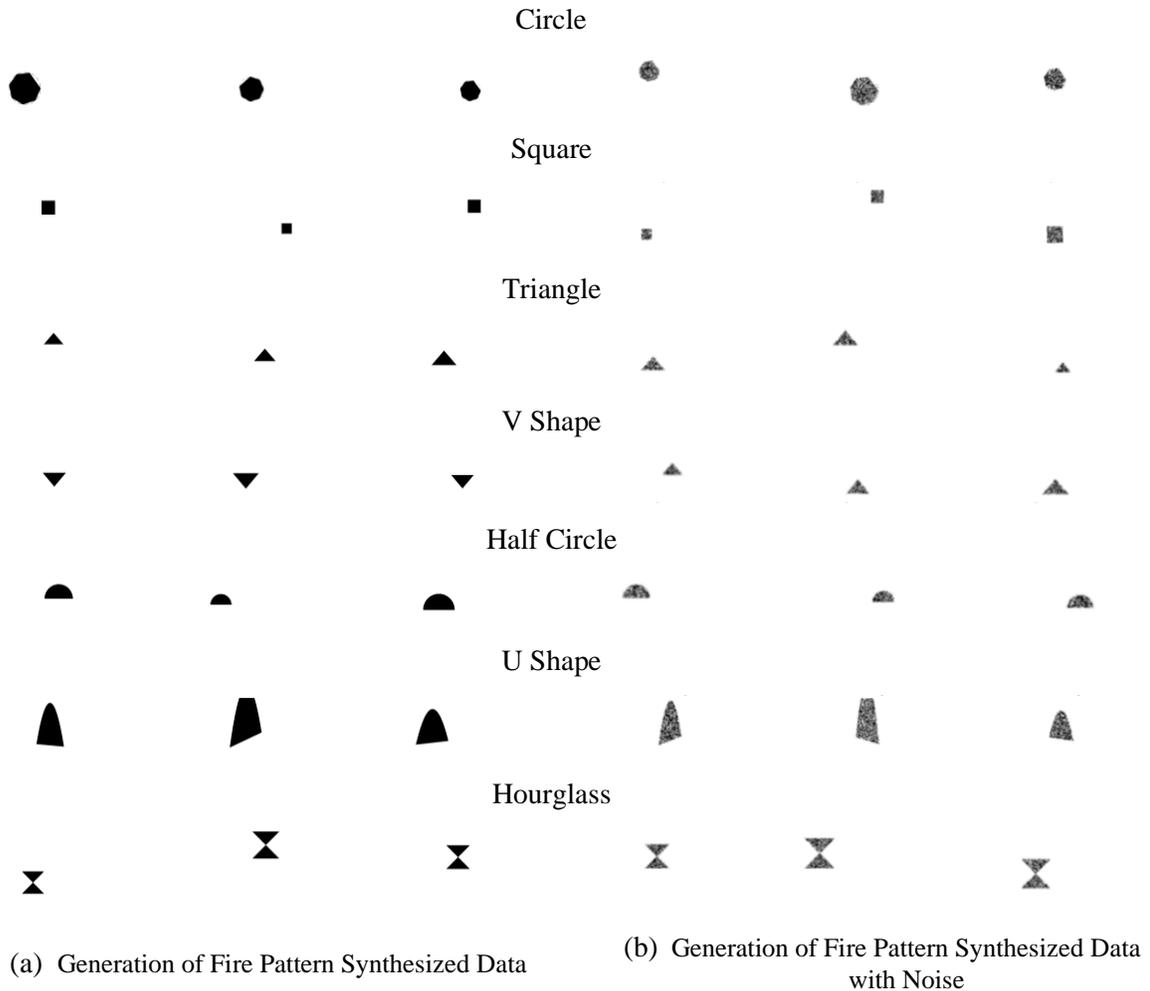

(a) Generation of Fire Pattern Synthesized Data

(b) Generation of Fire Pattern Synthesized Data with Noise

Figure. 7. Generation of fire pattern-synchronized data

### 5.2.2 Aspect Ratio Distributions of Synthetic Fire Patterns

Fig. 8 shows the aspect ratio distributions of the synthetic fire patterns. Despite the marked complexity and variability observed in the distribution, the overarching trend in the aspect ratio for each pattern consistently aligns with and is predictable based on the distinct geometric characteristics intrinsic to each fire pattern. Each shape is accompanied by a corresponding graph plotting the aspect ratio distribution as a function of the angle, suggesting a line segment's rotation around the shape's center. For the circle, the aspect ratio is a constant value, which is reflected in the flat line in the graph, indicating the stability of its aspect ratio regardless of the angle. The rectangle, triangles, and hourglass show repeating patterns in their aspect ratio graphs, characterized by peaks and valleys. This stability and predictability in aspect ratio trends are significant for synthesizing fire patterns. By understanding and replicating these stable aspect ratio distributions, algorithms can be trained to recognize and classify actual fire patterns based on the geometric features captured in the synthesized data. Therefore, these aspect ratios serve as the foundational training data for a model designed to classify the shape of a fire pattern by analyzing its aspect ratio distribution.

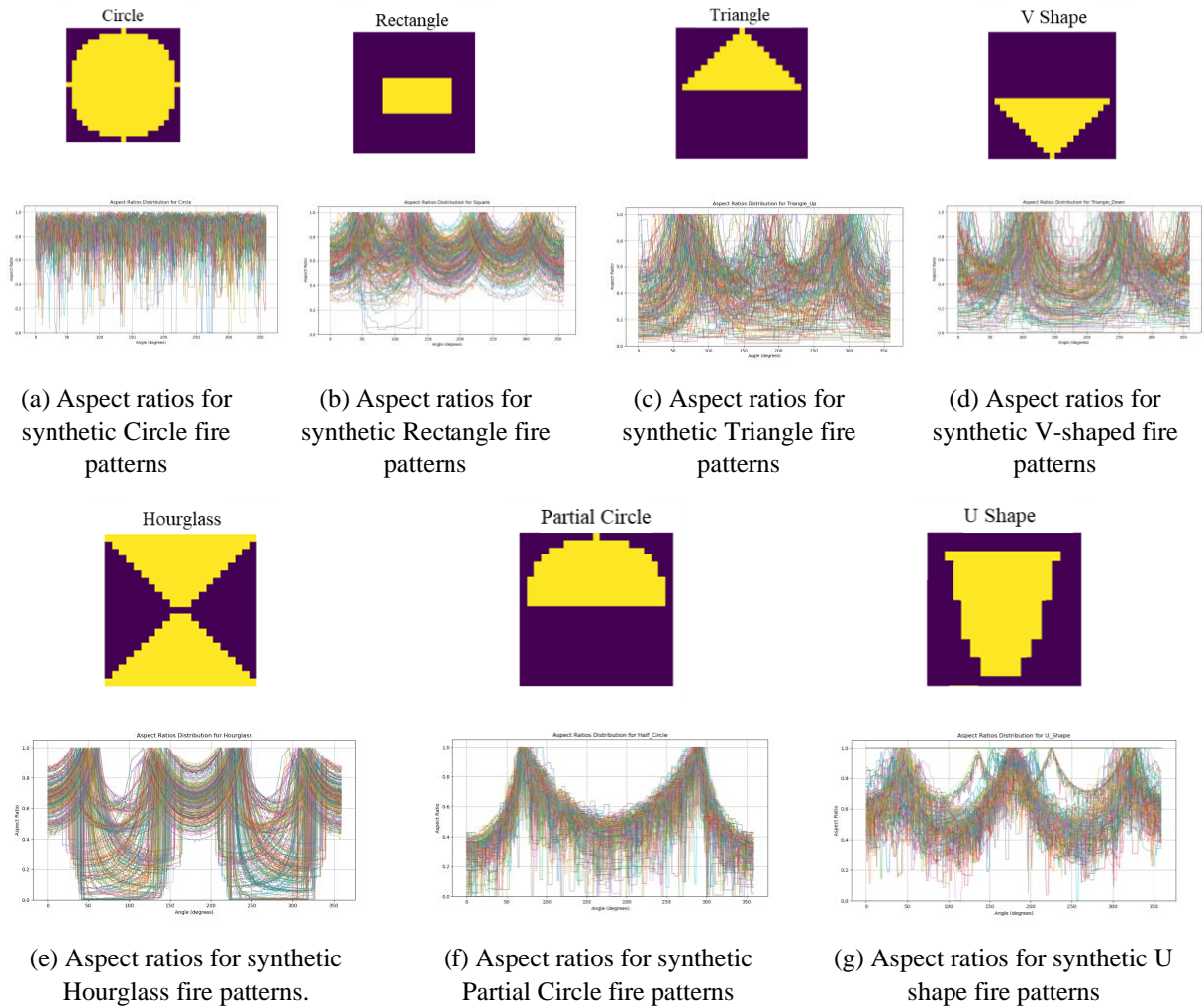

Figure. 8. Aspect ratio distributions for fire pattern-synchronized data

### 5.2.3 Aspect Ratio Distributions of Real Fire Patterns

We can apply this method to analyze real fire scenarios once we understand the aspect ratio distributions of standard fire patterns. For example, Fig. 9 illustrates the aspect ratio analysis based on fire patterns extracted from an indoor fire scene. Fig. 9 (a), (d), (g), (j), and (m) highlight the initial areas of interest within a red bounding box on a wall, which are crucial for fire pattern analysis. The segmented images shown in Fig. 9 (b), (e), (h), (k), and (n) show the fire patterns isolated digitally for aspect ratio analysis. The aspect ratios of the fire patterns from natural scenes exhibit a distribution pattern that, while somewhat irregular, generally aligns with that of synthetic triangle-down fire patterns. As shown in Fig. 9 (b), the triangle-up-shaped fire patterns feature three distinct peaks and two valleys. This specific configuration is mirrored in the synthetic models depicted in Fig. 9 (c). Moreover, the alignment of these peaks and valleys on the x-axis, representing the degree of rotation, further corroborates the similarity between natural and synthetic fire patterns. The close match between natural and synthetic patterns suggests that synthetic models are robust and can reliably replicate the critical characteristics of actual fire patterns.

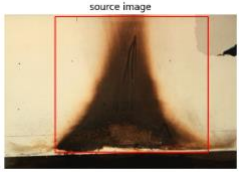 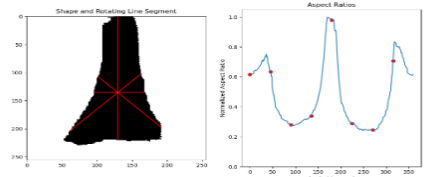 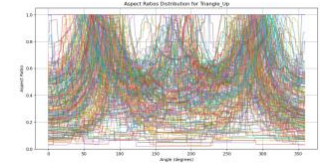

(a) Triangle fire patterns　　(b) Aspect ratios of triangle fire patterns　　(c) Aspect ratios for synthetic triangle fire patterns

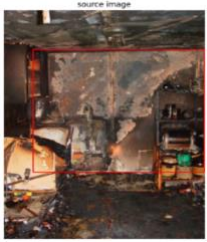 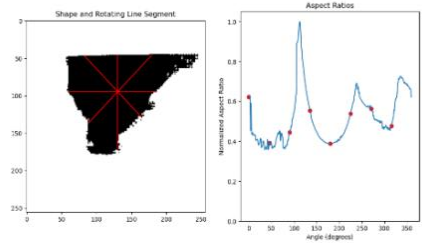 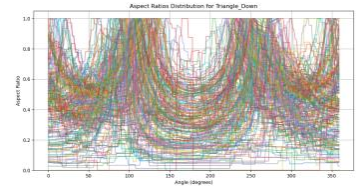

(d) V-shaped fire patterns　　(e) Aspect ratios of V-shaped fire patterns　　(f) Aspect ratios for synthetic V-shaped fire patterns

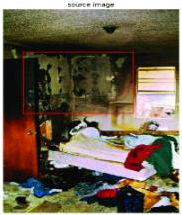 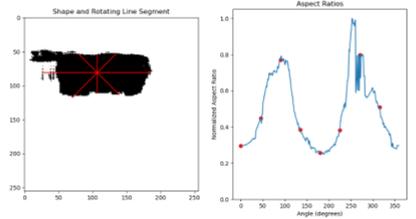 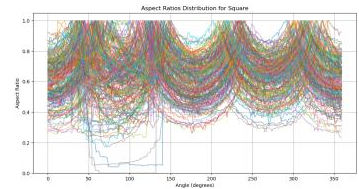

(g) Rectangular fire patterns　　(h) Aspect ratios of rectangular fire patterns　　(i) Aspect ratios for synthetic rectangular fire patterns

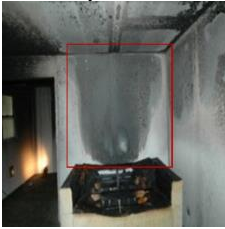 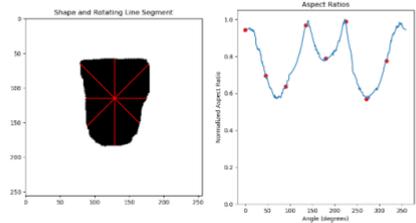 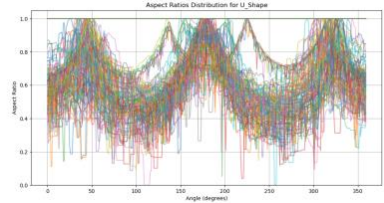

(j) U-shaped fire patterns　　(k) Aspect ratios of U-shaped fire patterns　　(l) Aspect ratios for synthetic U-shaped fire patterns

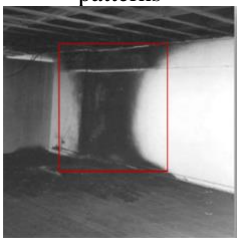 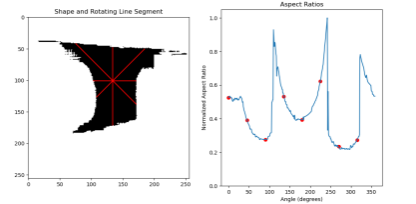 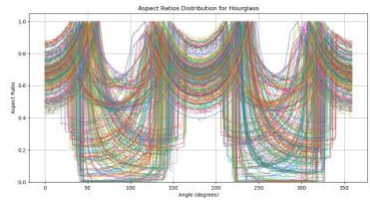

(m) Hourglass-shaped fire patterns　　(n) Aspect ratios of hourglass-shaped fire patterns　　(o) Aspect ratios for synthetic hourglass fire patterns

Figure. 9. Aspect ratio distributions of extracted fire patterns from real indoor fire scenes

## 5.3 Fire Pattern Classification Based on Aspect Ratios with Physical Features

### 5.3.1 Physical Features Definition

Quantitative assessment of fire patterns requires defining specific physical features that can be consistently quantified in various scenarios. These features include the number of peaks and valleys along the pattern's edge and their precise locations, measured in rotation degrees from a fixed reference point. This systematic approach allows objective comparison and classification of fire patterns based on their geometric characteristics. As discussed in Section 5.1, aspect ratios are utilized to quantify the fire patterns. Additionally, three related physical features based on aspect ratios are outlined: aspect ratio peaks, valleys, and rotation degrees. Peaks in the aspect ratio are points on the edge of a fire pattern that represents the maximum outward extension within a specific segment. Aspect ratio valleys are points denoting the maximum inward recession within a segment of a fire pattern. The angle of rotation is the angular measurement, in degrees, used to specify the locations of peaks and valleys. The reference point, set at 0 degrees, is typically the pattern's topmost point, with the rotation angle increasing in a counterclockwise direction. The definition and tabulation of these physical features are critical to the methodology that enhances the automation and accuracy of fire pattern analysis. Below is Table 1, which summarizes the data for various fire patterns, illustrating the numbers and locations (rotation degrees) of the peaks and valleys.

**Table 1. The number and locations (rotation angles) of peaks and valleys for fire patterns**

| Fire Patterns | Average Numbers of Peaks | Average Peak Locations (degrees) | Average Numbers of Valleys | Average Valley Locations (degrees) |
|---|---|---|---|---|
| Circle | 0 | 0 | 0 | 0 |
| Half-Circle | 2 | 76, 268 | 1 | 182 |
| Hourglass | 4 | 50, 135, 225, 337 | 3 | 76, 182, 278 |
| Rectangular | 4 | 52, 126, 226, 315 | 3 | 88, 177, 272 |
| Triangle | 3 | 67, 173, 293 | 2 | 129, 225 |
| V shape | 2 | 110, 249 | 3 | 45, 175, 316 |
| U shape | 3 | 72, 216, 293, 313 | 2 | 119, 270 |

### 5.3.2 Classification Model Based on Aspect Ratios with Physical Features

Based on predefined physical features, such as 360-dimensional aspect ratio vectors, numbers of peaks and valleys, and their locations (represented as 10-dimensional vectors), the random forest algorithm is trained using synthetic fire patterns and subsequently evaluated against actual fire patterns, as depicted in Fig. 10. The classification process begins when an input data point, representing a fire pattern, is introduced to the random forest. This sample undergoes processing through multiple decision trees within the forest, as illustrated by the branching structure beneath the new sample. Each decision tree within the random forest processes the sample independently and arrives at its result. At the end of this process, the results of each tree are aggregated using majority voting for classification tasks. Majority voting means that the most common outcome among all trees is selected as the final prediction for the classification of a fire pattern.

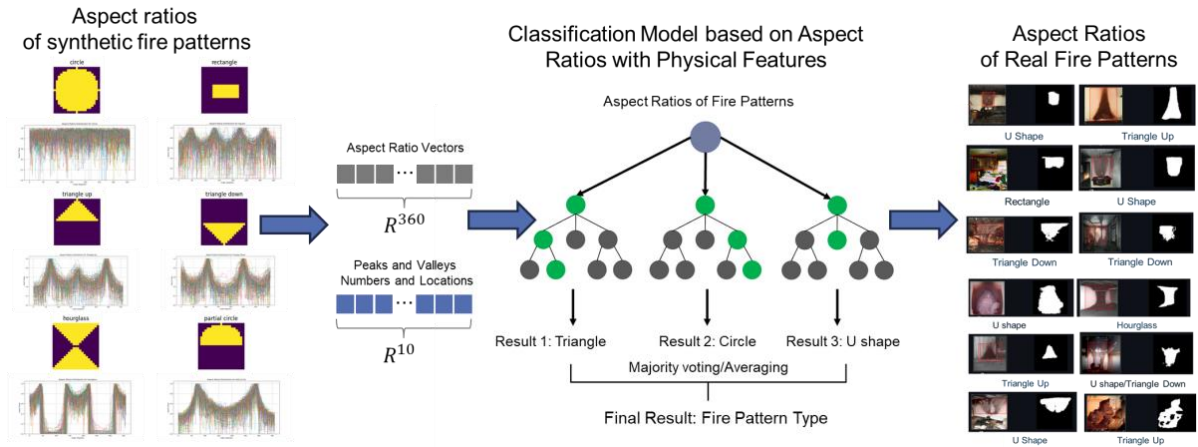

Figure. 10. Classification model based on Aspect ratios with physical characteristics

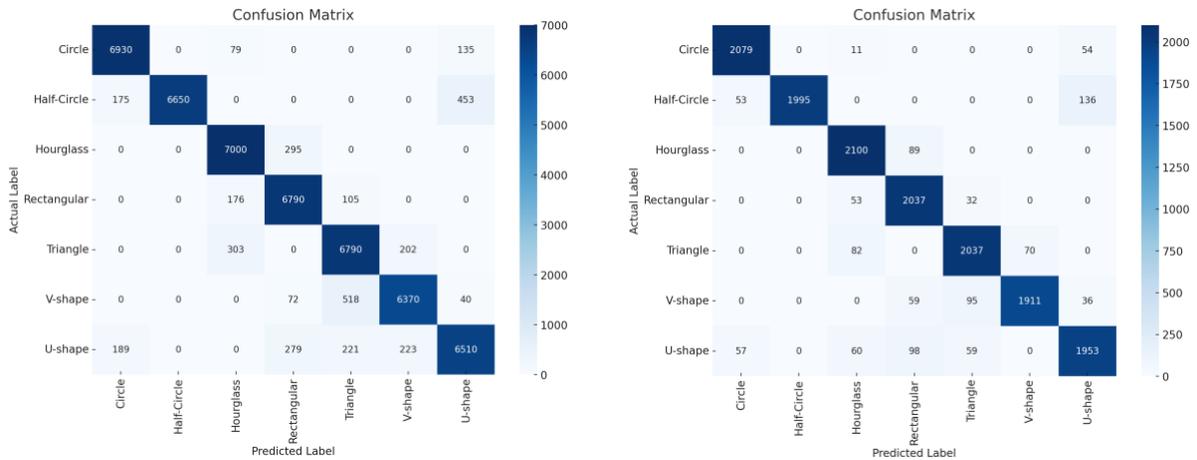

(a) Traning data confusion matrix  (b) Testing data confusion matrix
Figure 11. Classification performance of the synthetic fire patterns model

Fig. 11 shows two confusion matrices, one for training samples (70% of all data) and another for testing samples (30%). Both matrices indicate that the model accurately classifies different shapes with minimal false positives and false negatives. The first matrix on the left, representing the training data, shows the model's strong performance, with an overall accuracy of 0.93 (93%). Precision, recall, and F1 scores are consistently high across all classes, with several classes achieving near-perfect scores. There are a few misclassifications, as seen in the non-diagonal elements of the matrix. The second matrix on the right, representing the testing data, similarly demonstrates high accuracy at 0.93 (93%), with most classes maintaining excellent precision, recall, and F1 scores.

Table 2. Classification results in fire patterns from synthetic fire patterns

| Fire Patterns | Precision | Recall | F1-Score |
|---|---|---|---|
| Training | | | |
| Circle | 0.9501 | 0.9699 | 0.9598 |
| Half-Circle | 1 | 0.9137 | 0.9553 |
| Hourglass | 0.9261 | 0.9596 | 0.9436 |
| Rectangular | 0.9132 | 0.9603 | 0.936 |
| Triangle | 0.8894 | 0.9307 | 0.9087 |
| V shape | 0.9375 | 0.91 | 0.9233 |
| U shape | 0.912 | 0.8769 | 0.894 |
| Average | 0.9326 | 0.9316 | 0.9315 |
| Accuracy | | | 0.9313 |
| Testing | | | |
| Circle | 0.9498 | 0.9697 | 0.9596 |
| Half-Circle | 1 | 0.9135 | 0.9546 |
| Hourglass | 0.935 | 0.9593 | 0.947 |
| Rectangular | 0.9073 | 0.96 | 0.9326 |
| Triangle | 0.9006 | 0.9306 | 0.9163 |
| V shape | 0.9368 | 0.9095 | 0.9237 |
| U shape | 0.8963 | 0.8769 | 0.8871 |
| Average | 0.9323 | 0.9314 | 0.9316 |
| **Accuracy** | | | 0.9312 |

Fig. 12 illustrates the classification performance of the random forest model with physical features for various actual fire patterns. For the triangle-down fire pattern, as shown in Fig. 12 (a), (b), and (c), the model predicts it as triangle-down with a probability (confidence) of 73%, triangle-up with 23%, and rectangular with 4%, with no significant probabilities assigned to other shapes. In the case of the triangle-up fire pattern, as shown in Figs. 12 (d), (e), and (f), the model shows a strong preference for correctly classifying it as triangle-up at 85%, with smaller probabilities for 'Hourglass' at 9% and triangle-down at 5%. The U-shaped fire pattern, as shown in Fig.12 (g)(h)(i), is predominantly recognized as a U-shape at 75%, followed by rectangular at 10%, and minimal probabilities for circle at 4%, triangle-down and triangle-up each at 4%.

| | | |
|---|---|---|
| 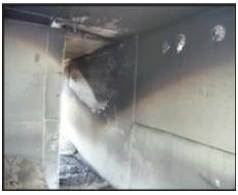 | 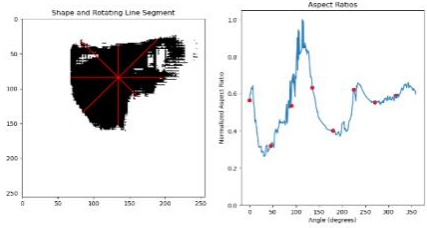 | 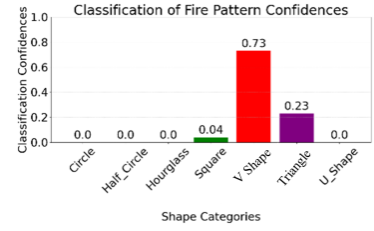 |
| (a) V-shaped fire patterns | (b) Extracted V-shaped fire patterns and related aspect ratios | (c) Classification Confidence of the V-shaped Fire Patterns |
| 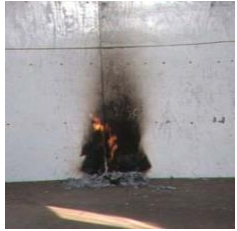 | 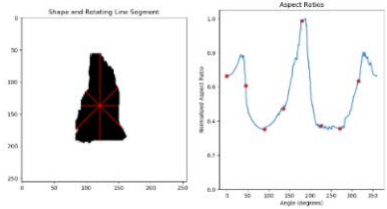 | 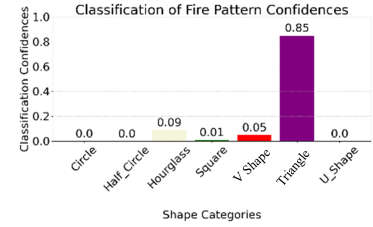 |
| (d) Triangle Fire Patterns | (e) Extracted triangle fire patterns and related aspect ratios | (f) Classification Confidence of the Triangle Fire Patterns |
| 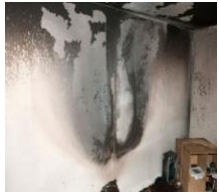 | 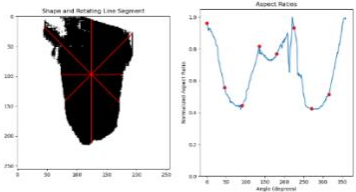 | 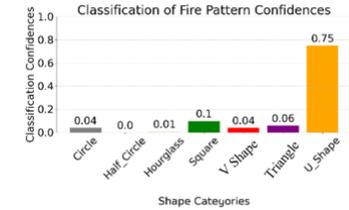 |
| (g) U-shaped fire patterns | (h) Extracted U-shaped fire patterns and related aspect ratios | (i) Classification Confidence of the U-shaped Fire Patterns |

Figure 12. Model Classification Performance of Natural Fire Patterns

Table 3. Test results on the fire patterns from actual fire scenes

| Methods | Fire Patterns | Precision | Recall | F1 score |
|---|---|---|---|---|
| Random Forest with Aspect Ratios | Circle | 0.75 | 1.00 | 0.86 |
| | Half Circle | 1 | 0.5 | 0.67 |
| | Hourglass | 0.8 | 1.00 | 0.89 |
| | Rectangular | 0.75 | 1.00 | 0.86 |
| | Triangle | 0.67 | 0.5 | 0.57 |
| | V-shape | 1.00 | 1.00 | 1.00 |
| | U-Shape | 1.00 | 0.75 | 0.86 |
| | Average | 0.85 | 0.83 | 0.82 |
| | Accuracy | | | 0.83 |
| Convolutional neural networks (CNN) | Circle | 0.50 | 0.83 | 0.62 |
| | Half Circle | 0 | 0 | 0 |
| | Hourglass | 0.50 | 1.00 | 0.67 |
| | Rectangular | 1 | 0.67 | 0.8 |
| | Triangle | 0.40 | 0.50 | 0.44 |
| | V-shape | 0.33 | 0.33 | 0.33 |
| | U-Shape | 0.75 | 0.38 | 0.50 |
| | Average | 0.56 | 0.52 | 0.50 |
| | Accuracy | | | 0.52 |

Table 3 presents the test results for different fire patterns using precision, recall, and F1-Score as evaluation metrics. The evaluated fire patterns include circles, hourglasses, rectangular, triangles down, triangles up, and U shapes. Notably, the circle pattern resulted in scores of 0.00 for all metrics due to the absence of actual fire patterns for this shape. The model achieved perfect scores (precision, recall, and F1-Score of 1.00) for the Hourglass, Rectangular, and Triangle Up patterns. For the Triangle Down pattern, the model recorded a precision of 0.60, a recall of 0.75, and an F1 score of 0.67, indicating moderate performance. The U shape pattern showed a precision of 1.00, a recall of 0.65, and an F1-Score of 0.79, reflecting good but imperfect performance. Generally, the averages for precision, recall, and F1-Score in all patterns were 0.77, 0.73, and 0.74, respectively, with an overall accuracy of 0.79. The Convolutional Neural Network (CNN) approach achieved lower performance, with an average precision of 0.48, recall of 0.45, F1 score of 0.44, and an overall accuracy of 0.53. This demonstrates the superior effectiveness of the Random Forest approach for fire pattern classification in this application.

### 5.3.3 Explanation of the Classification Model

Fig. 13 (a) provides a visualization of the simplified decision tree used to classify various fire patterns based on aspect ratios at different rotation angles. The figure illustrates the branching of the decision tree at different rotation angle thresholds. For example, at the rotation angle of 76 degrees, the tree splits into two branches based on the aspect ratio threshold of 0.676. Further divisions occur at 182 degrees and 337 degrees based on the aspect ratio thresholds of 0.5132 and 0.5694, respectively. Subsequent splits occur at 338 degrees, 9 degrees, 266 degrees, and 20 degrees. Each branch represents a decision node that further classifies the patterns based on their rotation angles. The bottom section features pie charts that visualize the proportion of each fire pattern within the specific rotation degree clusters determined by the decision tree. Each pie chart at the bottom indicates the number of occurrences (n) for each pattern within a particular range of rotation angles. For example, at 337 degrees, the pie chart shows 222 instances of the Triangle Down pattern and 40 cases of the Hourglass pattern. Overall, Fig. 13 (a) demonstrates how the decision tree effectively learns to classify fire patterns by splitting at specific rotation angles and aspect ratio thresholds, categorizing patterns into predefined groups. Figs. 13 (b)-(d) illustrate scatter plots showing the separation of various fire patterns at specific rotation angles (76 degrees, 182 degrees, and 337 degrees). Each plot helps visualize how fire patterns are distributed and categorized along specific thresholds, highlighting the effectiveness of the decision tree in classifying these patterns based on rotation angles and aspect ratios. For example, Fig. 14 (b) shows how patterns are divided based on the aspect ratio threshold of 0.676 at 76 degrees. For instance, patterns such as U shape, Hourglass, and Triangle-down are predominantly found below the threshold, while patterns like Circle, Half Circle, and Rectangular are mostly above it, indicating clear separation based on the aspect ratio. The detailed explanations of the classification path of the decision tree are illustrated in Fig. 14. The decision tree can classify the fire patterns through a series of steps. For example, to classify U-shaped fire patterns, the decision tree follows these steps: (1) Evaluate if the aspect ratio at 76 degrees is less than 0.676. (2) Check if the aspect ratio at 76 degrees is more significant than 0.5132. (3) Determine if the aspect ratio at 9 degrees is more significant than 0.7542. These steps, as depicted in Fig. 14(c), demonstrate how the decision tree uses specific criteria based on aspect ratios at different rotation degrees to classify fire patterns accurately.

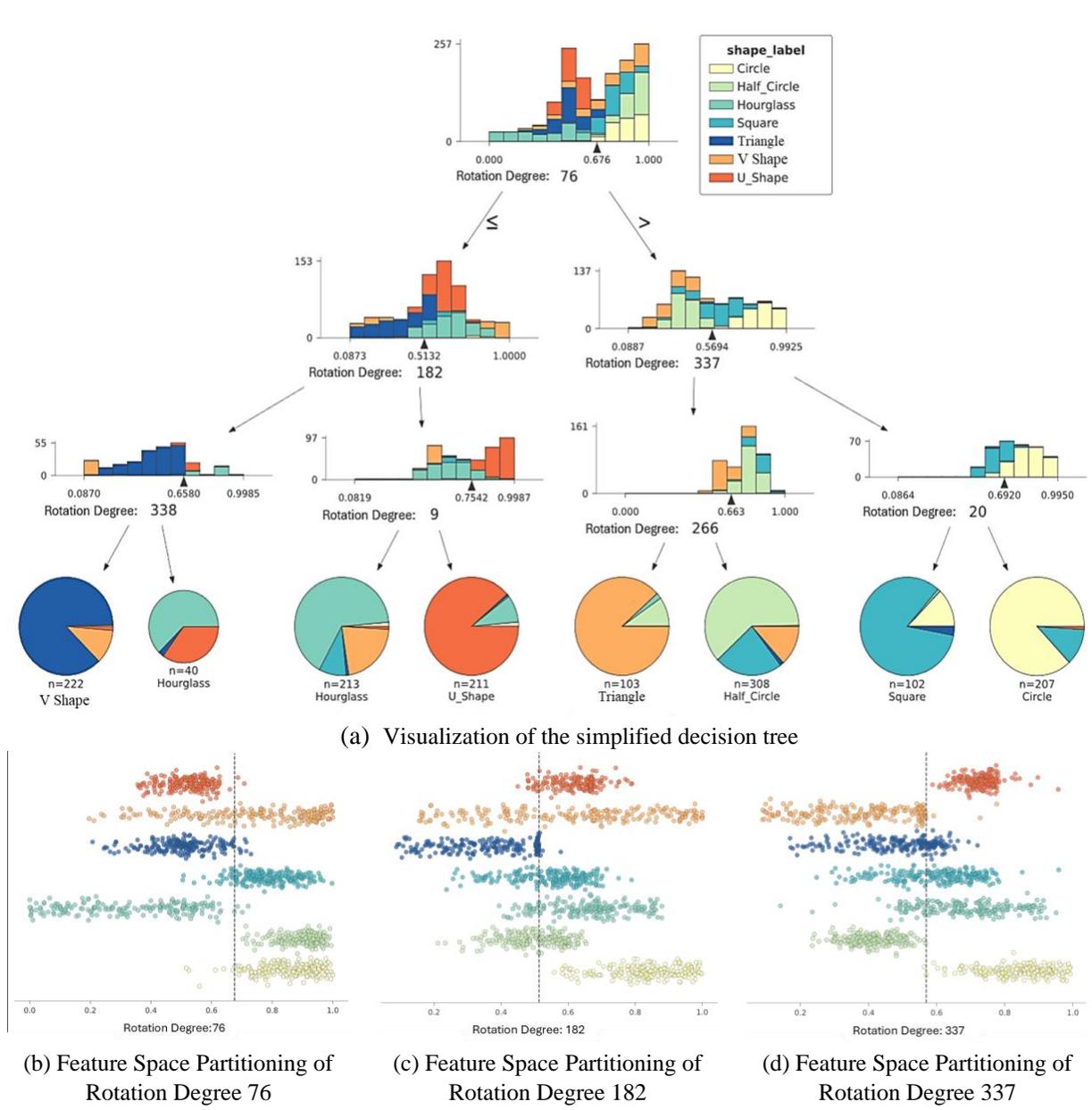

(a) Visualization of the simplified decision tree

(b) Feature Space Partitioning of Rotation Degree 76

(c) Feature Space Partitioning of Rotation Degree 182

(d) Feature Space Partitioning of Rotation Degree 337

Figure 13. Visualization of the decision tree structures

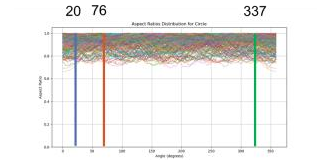
(a) Aspect ratios of Circle fire patterns

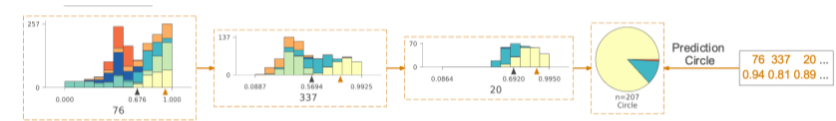
(b) Classification path of the Circle fire patterns

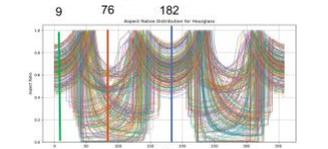
(c) Aspect ratios of U-shaped fire patterns

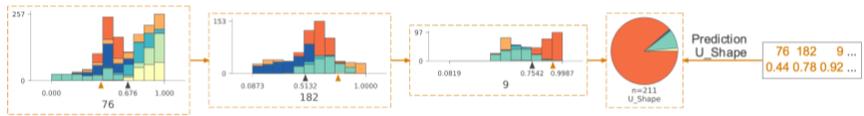
(d) Classification path of the U-shaped fire patterns

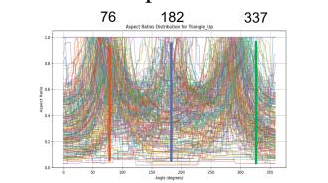
(e) Aspect ratios of Triangle fire patterns

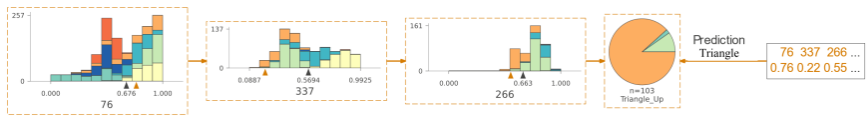
(f) Classification path of the Triangle fire patterns

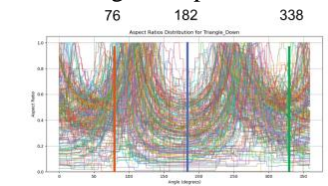
(g) Aspect ratios of V-shaped fire patterns

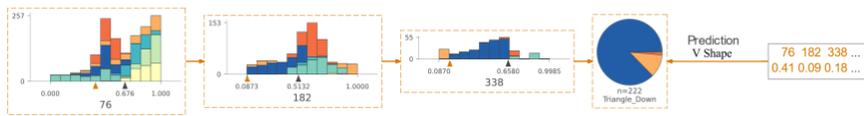
(h) Classification path of V-shaped fire patterns

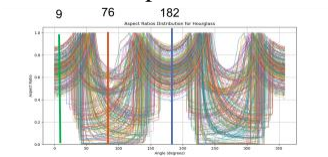
(i) Aspect ratios of Hourglass fire patterns

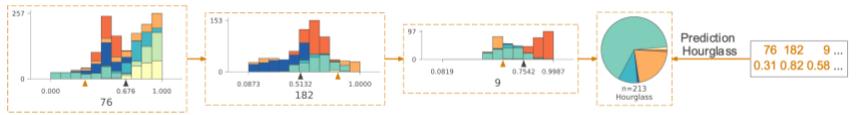
(j) Classification path of the Hourglass fire patterns

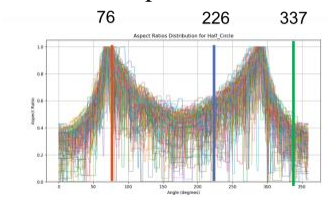
(k) Aspect ratios of Half-Circle fire patterns

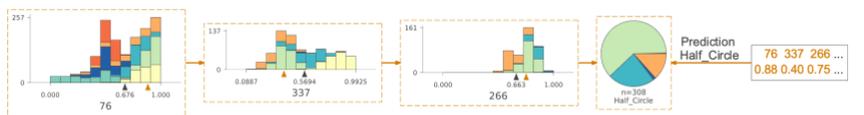
(l) Classification path of the Half-Circle fire patterns

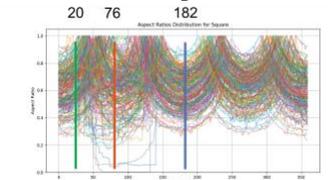
(m) Aspect ratios of Rectangular fire patterns

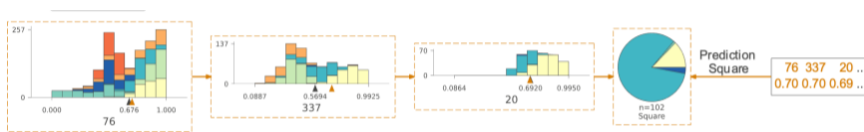
(n) Classification path of Rectangular fire patterns

Figure 14. Classification Path Explanations of the Decision Tree

# 6. Segmentation of Point Clouds for Spatial Relationship of Fire Scenes

## 6.1 Task Definition for Fire Scene Segmentation

As shown in Fig. 15, the image presents an indoor fire scene generated from point-cloud data. This comprehensive model accentuates vital structural features such as walls, floors, ceilings, and furniture. Notable variations in coloration and texture on different surfaces suggest differential exposure to heat and smoke, characteristic of the aftermath of a fire. In the context of fire pattern analysis, the objective is to apply segmentation techniques to delineate and classify distinct fire patterns on various walls. In addition, establishing a correlation between these patterns and adjacent objects is crucial. Such associations are instrumental for fire investigators in reconstructing the sequence of events, understanding the progression of the blaze, and ultimately determining fire origin.

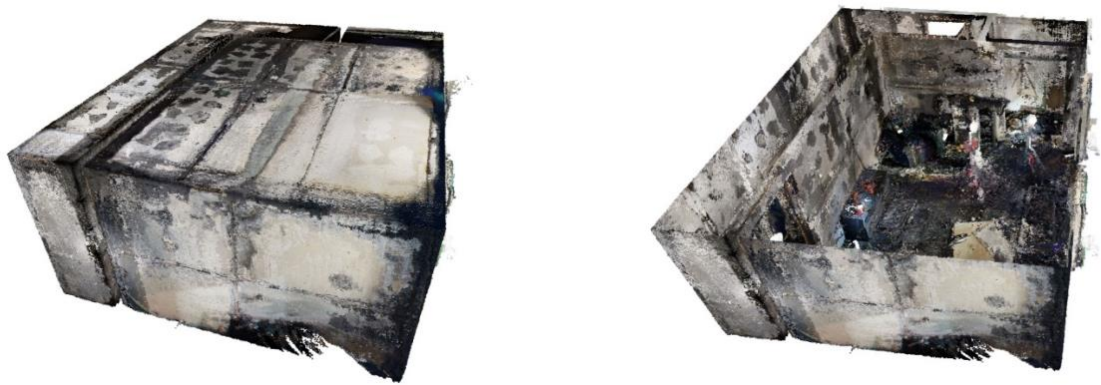

(a) Outdoor view  (b) Indoor view
Figure 15. Illustrations of an indoor 3D scene constructed from point clouds

## 6.2 Instance Segmentation for Indoor Scenes

Point-cloud segmentation is instrumental in parsing and classifying the constituent elements of a point cloud, a process that is especially pertinent in the analysis of indoor fire scenes. This segmentation is divided into two principal types: semantic and instance segmentation. Semantic Segmentation: This process involves categorizing each point within a point cloud into discrete classes according to inherent attributes or features. This categorization is typically aligned with the physical elements of a room, delineating walls, floors, doors, and windows. This level of segmentation is fundamental for the interpretation of the physical layout and structure within 3D spatial models, facilitating a comprehensive understanding of the space under consideration. Instance Segmentation: This method extends beyond the categorical differentiation semantic segmentation offers. Instance segmentation discerns discrete occurrences of categorized objects, enabling differentiation between multiple instances of the same class. For example, within an indoor setting, this segmentation identifies structural elements as walls and distinguishes one wall from another. Consequently, this facilitates nuanced and detailed modeling of each element within the scene, which is paramount for identifying individual fire patterns, such as those occurring on different walls, and associating these patterns with corresponding indoor objects. The precision afforded by instance segmentation is particularly vital in fire scene investigations. It allows for identifying and classifying fire patterns and affected objects, providing crucial insights into the origin and progression of a fire.

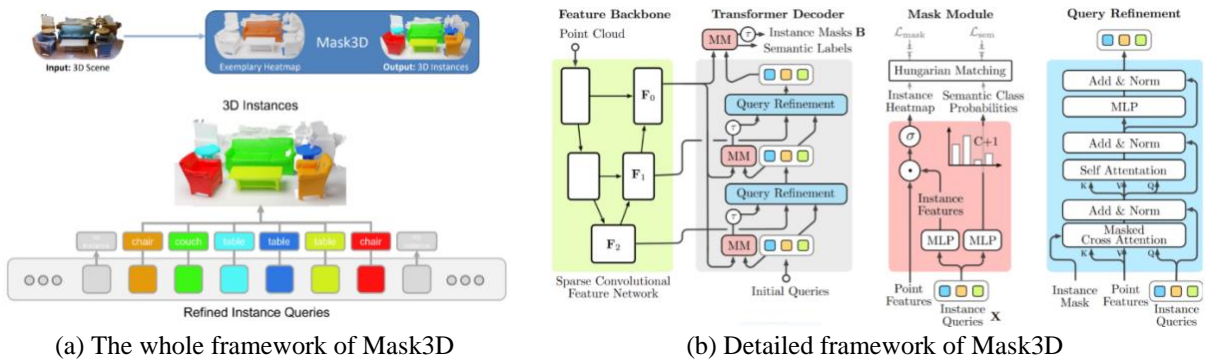

(a) The whole framework of Mask3D    (b) Detailed framework of Mask3D

Figure 16. Transformer-based model for 3D point cloud instance segmentation [34]

Fig. 16 shows the Mask3D model for 3D instance segmentation [34]. Integrates a feature backbone and a Transformer decoder. The model's centerpiece is the instance queries, representing individual object instances and predicting point-level masks. These queries undergo iterative refinement through the Transformer decoder, facilitating cross-attention to point features from the backbone and self-attention among queries. Repeated across multiple iterations and scales, this process leads to refined instance queries. The mask module processes these queries with point features to output a semantic class and binary mask for each query. Key features include (1) Instance Queries: Using learned instance queries to segment objects, eliminating the need for heuristic design choices. (2) Transformer Decoders: Employ multiple Transformer decoders to refine these queries by attending to multiscale point cloud features. (3) Direct Mask Prediction: Directly generates instance masks from point clouds, avoiding traditional voting or clustering. (4) Sparse Convolution Backbone: Utilizes a backbone network for effective feature extraction from sparse 3D data.

As shown in Fig. 17. (a), the segmentation algorithm has isolated the main elements such as walls, floors, and ceilings. These are color-coded and separated from the overall model to demonstrate the algorithm's capacity to recognize and categorize the core constructs of the building unit. Different colors denote separate instances, individual walls, flooring sections, and ceiling areas, providing clarity on spatial layout and architectural boundaries. Fig. 18 (b) shows the segmentation of movable objects within the room, including chairs, a bookcase, a cabinet, and a sofa. Each piece of furniture is encircled and labeled, demonstrating the algorithm's ability to distinguish between diverse object types and their instances within the context of the room.

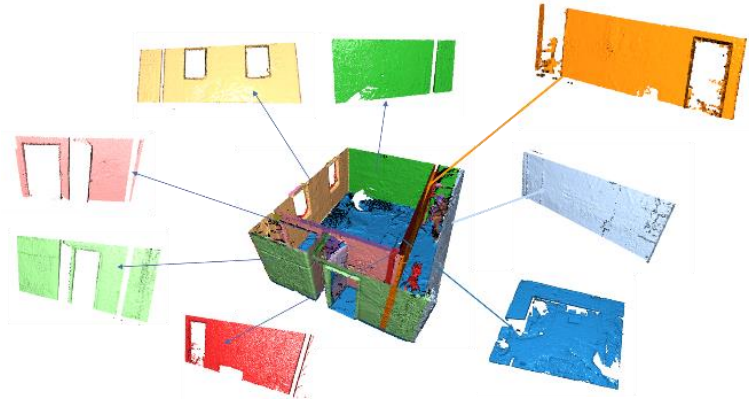

(a) Segmentation results of structural component

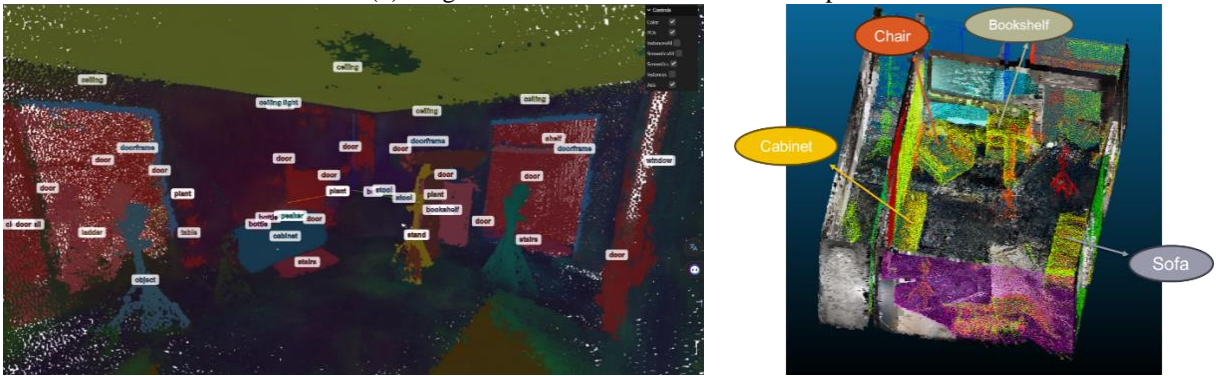

(b) Results of the Segmentation of Indoor Furniture

Figure 17. Multidimensional Fire Pattern Mapping Process

## 6.3 Spatial Relationship Mapping and Analysis for Indoor Fire Scenes

To better understand the spatial relationships in indoor fire scenes, we integrate 2D fire scene images processed using the SAM model with 3D point cloud data segmented by the Mask3D model. The process begins by segmenting the 3D point cloud data to identify structural components, which are then projected onto 2D fire scene images to extract fire patterns. These patterns are carefully classified and projected back onto the 3D model, ensuring accurate alignment with the spatial geometry of the scene. Fig.18 illustrates a comprehensive procedure for mapping 2D fire patterns onto a 3D reconstruction of a fire-affected scene. This vital process synthesizes planar image data with the 3D spatial environments to enhance the analysis of fire patterns. After the 3D segmentation, architectural components such as walls, floors, and ceilings are isolated. The segmented 3D components are then mapped to 2D images to obtain visual representations of walls with fire patterns. 2D images undergo a detailed analysis through SAM to identify fire patterns. These identified 2D fire patterns are then projected onto the corresponding 3D structures, integrating the detailed textural information from 2D images with the spatial data of the 3D model. The projected patterns are meticulously aligned with the 3D model, ensuring that the patterns adhere to the contours and geometry of the actual space. This step is critical to maintaining spatial coherence between the two data sets. In the final visualization, distinct fire patterns are labeled and highlighted in the 3D model, revealing the extent and specific location of fire damage on different surfaces and objects within the room.

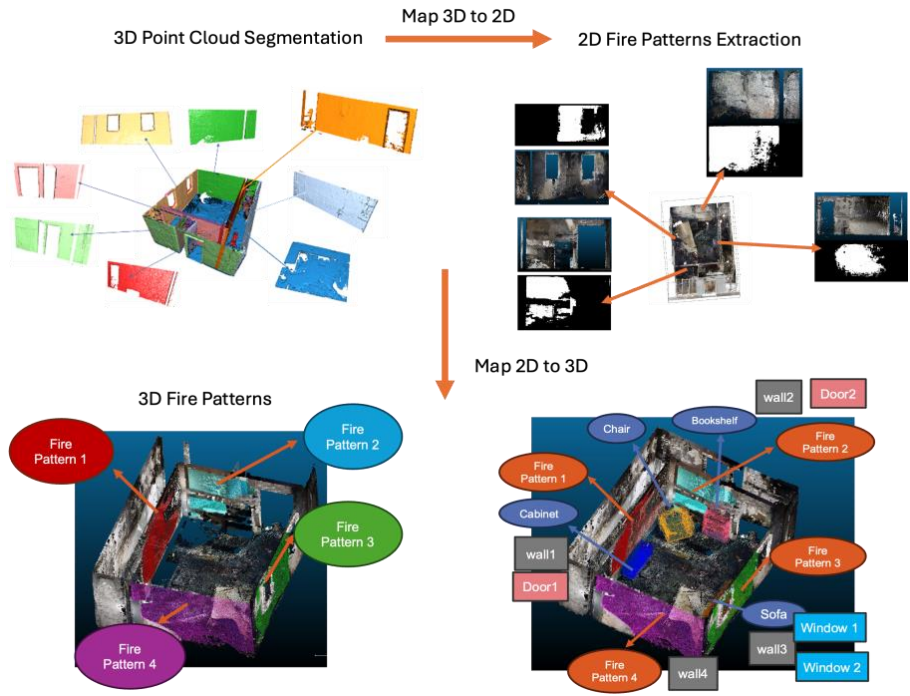

Figure 18. Multidimensional Fire Pattern Mapping Process

One of the prominent limitations in contemporary 3D segmentation algorithms is their inability to accurately extract fire patterns on walls. Although these conventional approaches are proficient at delineating the essential structural elements and objects within an environment, they often fail to capture fire patterns' intricate and nuanced manifestations accurately. This deficiency can lead to significant errors in analyzing the origin and evolution of a fire event. To overcome this challenge, our work introduces an innovative, integrated method that combines 2D fire pattern extraction, utilizing the SAM detailed in Section 4, with 3D point cloud segmentation of fire scenes, as discussed in Section 5. This synthesis facilitates the construction of a detailed 3D fire scene graph. Upon the segmentation of fire patterns in 2D, our methodology adeptly projects and aligns these segments onto the corresponding three-dimensional environment. This approach confers several advantages: Firstly, it uses enhanced detail and clarity provided by 2D imagery to detect the shapes of fire patterns with heightened accuracy. Second, projecting these precise segments into a 3D segmentation yields a more holistic and spatially conscious depiction of the fire-impacted zones, as shown in Algorithm 1.

By separating these components, the scene can be meticulously analyzed to detect fire patterns from each separated surface and allow the detected fire patterns from 2D images back to the interior surfaces of a 3D fire scene to help understand the fire patterns spatial relationship as shown in Fig. 18. Distinct fire patterns are labeled and highlighted in the 3D model, showing the extent and specific locations of fire damage across surfaces and objects within the room. By isolating these components, the scene can be meticulously analyzed, enabling the detection of fire patterns on individual surfaces. These patterns are then mapped back onto the interior surfaces of the 3D fire scene, enhancing the understanding of the spatial relationships between them, as demonstrated in Fig. 18.

| Algorithm 1. Automated Fire Scenes Spatial Relationship Analysis Algorithm |
|---|
| **Function:**<br>This algorithm integrates 2D fire pattern extraction with 3D scene segmentation using the SAM and a 3D segmentation model (Mask3D) to enrich fire pattern spatial relationship analysis.<br>**Input:**<br>- $I_{2D}$: 2D images of fire scenes processed by the SAM model.<br>- $P_{3D}$: 3D point cloud data segmentation processed by the Mask3D model.<br>**Output:**<br>A comprehensive scene graph $KG$ represents the segmented and classified fire patterns within the 3D fire scene.<br>1. **3D Point Cloud Segmentation**:<br>The 3D point cloud data $P_{3D}$ of the indoor scenes were segmented using the Mask3D model to distinguish structural components.<br>$$S_{3D} = Mask3D(P_{3D})$$<br>2. **Map 3D to 2D to Get Fire Patten Images**:<br>Project instances of walls and other components from the 3D segmentation $S_{3D}$ onto 2D fire pattern images $I_{2D}$ to establish a reference framework for pattern extraction.<br>$$M_{3D \to 2D} = Project(S_{3D}, I_{2D})$$<br>3. **2D Fire Pattern Extraction:**<br>Extract fire patterns from 2D images of walls using the SAM model, which isolates and classifies patterns based on visual evidence of fire damage.<br>$$F_{2D} = SAM(I_{2D})$$<br>4. **Map 2D to 3D**:<br>Overlay the segmented 2D fire patterns $F_{2D}$ onto the 3D model, employing projection algorithms to translate the pattern data into the spatial domain of the 3D scene.<br>$$F_{3D} = Project(F_{2D}, S_{3D})$$<br>5. **Refine Alignment:**<br>Fine-tune the alignment of the 2D fire pattern $F_{3D}$ projections on the 3D model, ensuring a precise placement that is congruent with the contours and geometry of the 3D space.<br>$$F'_{3D} = RefineAlignment(F_{3D}, S_{3D})$$<br>6. **Distance Calculation:**<br>To quantify their spatial relationships and influence, compute the distances between different fire patterns, structural elements, and furniture in the 3D space.<br>$$D = \{d_{ij} = \| x_i - x_j \| \mid x_i, x_j \in F'_{3D} \cup S_{3D}\}$$<br>**End of Algorithm**<br><br>**Notation:**<br>$I_{2D}$: 2D images of the fire scene.<br>$P_{3D}$: 3D point cloud data.<br>$S_{3D}$: Segmented 3D structural components.<br>$M_{3D \to 2D}$: Mapped 3D to 2D reference framework.<br>$F_{2D}$: 2D fire patterns extracted.<br>$F_{3D}$: Projected 2D fire patterns onto a 3D model.<br>$F'_{3D}$: Refined 3D fire patterns.<br>$D$: Distances between fire patterns, structural elements, and furniture. |

# 7. Discussion and Future Research

In this section, we discuss key aspects that are critical for advancing fire scene analysis, particularly in understanding the limitations of current methodologies and exploring future research possibilities. The discussions in Sections 7.1 and 7.2 are essential for addressing both the challenges of accurately capturing fire patterns and the potential of artificial intelligence in enhancing spatial relationship analysis. These sections aim to provide a foundation for improving fire scene understanding through AI-driven approaches, scene graph generation, and spatial data interpretation.

## 7.1 Fire Scene Understanding through Scene Graph Generation

Building upon the spatial relationship mapping discussed in Section 6.3, creating a scene graph allows us to represent and analyze the intricate spatial dependencies in a fire-affected environment. By accurately mapping the geometric relationships between fire patterns, structural components, and objects in the scene, we can develop a comprehensive view of the fire scene and better understand the fire's progression and spread.

Section 6.3 discussed how fire patterns are extracted from 2D images using the SAM model and aligned with 3D structural elements through the Mask3D model. This process generates detailed spatial relationships between fire-affected surfaces, architectural features, and furniture. Once the distances and geometric relationships between these elements are calculated, they form the foundation for a scene graph, as shown in Fig. 19. The scene graph encapsulates the entire fire scene, representing each structural element, object, and fire pattern as nodes, with edges defining the spatial relationships between them. This graph-based representation provides a powerful tool for fire investigators, allowing them to visualize the room's layout, the spread of fire, and the interactions between various elements in the space. By understanding these relationships, we can gain critical insights into the fire's origin, its environmental evolution, and potential fire spread pathways.

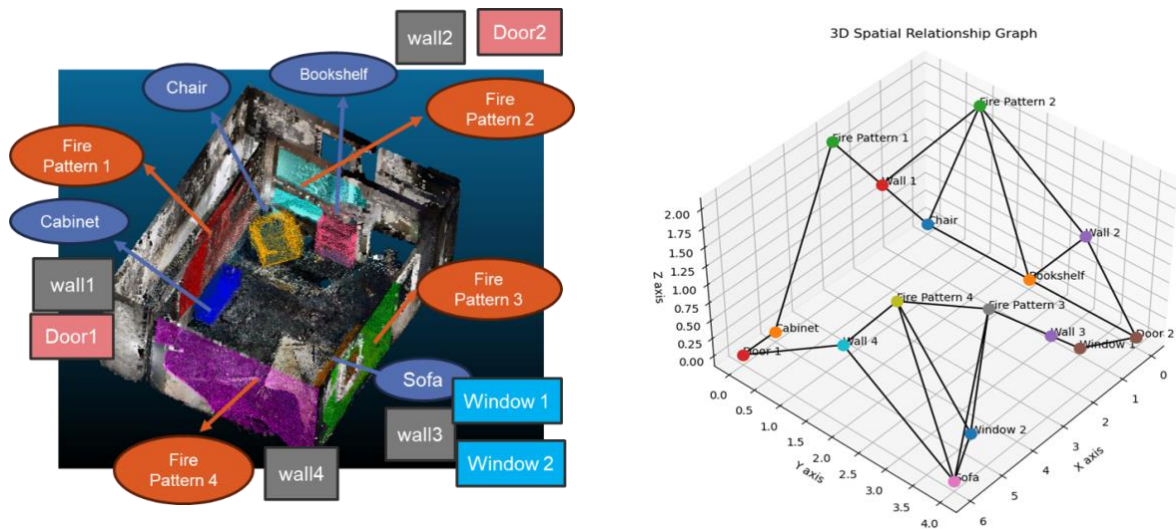

(a) Understanding of the indoor fire scene        (b) 3D spatial relationship graph

Figure 19. Understanding of the Indoor Fire Scene through Scene Graphs

## 7.2 Future Research

Future fire pattern recognition and spatial analysis research can greatly benefit from advancements in artificial intelligence and automated processes. Fire investigators typically examine burn patterns on various materials and surfaces to understand the development of a fire, which plays a critical role in determining its origin and cause.

(1) **Fire Pattern Recognition and Extraction**: Traditionally, fire investigators rely on visual inspections, either on-site or through photographs, to identify fire patterns. This manual process is not only time-consuming but also prone to human error, which can compromise the accuracy of the investigation. A promising research direction involves the integration of AI-powered image recognition algorithms that can automatically detect and highlight fire patterns in both digital images and 3D models. Through user-friendly interfaces such as touchscreens, voice commands, or graphical inputs, investigators can interact with the system to refine pattern recognition processes. This interactive capability allows the AI system to improve its accuracy over time, reducing the risk of human oversight and enhancing the speed and precision of fire pattern extraction.

(2) **Quantitative Definition and Classification of Fire Patterns**: Currently, the classification of fire patterns depends heavily on the experience and expertise of investigators, who categorize these patterns based on observed shapes and characteristics. Automating this process using AI could lead to significant improvements. By analyzing geometric shapes and providing consistent categorization, AI can help standardize the classification of fire patterns, offering a more reliable and data-driven approach to understanding fire behavior and progression. Such advancements will not only assist less-experienced investigators but also enhance the overall consistency and reliability of fire scene analyses.

(3) **Fire Pattern Spatial Relationship Analysis**: Another key area for future research is the spatial analysis of fire patterns in 3D environments. Investigators currently reconstruct fire scenes manually, creating 3D models or simulations to visualize how the fire progressed. This process, however, is labor-intensive and prone to inaccuracies. AI can greatly enhance the efficiency and accuracy of scene reconstruction. By processing 3D point cloud data, AI algorithms can automatically segment and isolate fire-affected areas, allowing for a detailed and accurate breakdown of fire patterns and damage within the scene. This speeds up the investigation and provides a more comprehensive view of the spatial relationships between fire damage and surrounding objects or structures.

(4) **Scene Graphs and Spatial Dependencies**: Understanding the dependencies and relationships between fire patterns and environmental factors is critical for reconstructing a fire's origin and behavior. Traditionally, investigators analyze these relationships manually, a process that can be complex and error-prone. AI-generated scene graphs, which map the connections between different fire patterns, structural components, and objects, can offer a more efficient and accurate method. These graphs will help investigators visualize potential fire spread pathways and dynamics, providing a deeper understanding of how fire interacts with its surroundings in both time and space. This area of research could significantly enhance the ability to predict fire behavior in complex environments.

By pursuing these directions, future research can leverage AI and advanced computing techniques to streamline fire investigations, reduce human error, and ultimately improve the accuracy and efficiency of fire pattern analysis and scene understanding.

## 8. Conclusions

Traditional methods for estimating surface fire damage heavily rely on fire investigators' visual assessments and experiential judgment, causing conflict and unreliable investigation results. Furthermore, relying solely on data-driven methods to classify fire patterns is problematic due to a fire scene's complexity and the limited data typically available. Integrating domain-specific physical knowledge into these methods is crucial to overcome these limitations and enhance the explainability and reliability of the resultant classifications. Therefore, we introduced an automated image-based framework for interpreting fire patterns that incorporate consistent quantitative assessments of fire pattern shapes and spatial relationships in the indoor fire scene. This framework improves the objectivity of fire pattern classifications, eliminating the bias inherent in human judgment. It supports fire pattern classification and location with several innovative contributions:

(1) Consistent Quantitative Shape Assessment: This approach standardizes the classification process by quantitatively defining fire patterns through aspect ratios, reducing the subjectivity associated with manual interpretations. The precision of this classification is bolstered by sophisticated algorithms that analyze geometrical and spatial characteristics, achieving 93% precision in the classification of synthetic fire patterns and 83% precision in the actual fire patterns, demonstrating the effectiveness and robustness of the proposed approach.

(2) Spatial relationship analyses and scene graph generation: The model's ability to evaluate the spatial relationships among different classified fire patterns and between fire patterns and the other objects in a scene aids in constructing a more detailed and accurate map of the fire scene, providing an interpretable representation of the evidence in a scene and thus supporting fire origin determination.

Despite these advances, several limitations must be addressed: (1) Complexity in Fire Origin Determination: Determining a fire's origin remains a complex challenge that necessitates integrating more comprehensive physical knowledge and other types of evidence (rather than just fire patterns) into our models. Expanding the scene graphs to include various variables and scenarios will enhance the interpretative capabilities for determining fire origins, improving the applicability and accuracy of the model. (2) Limited case studies: Our framework has been tested with limited fire scenes. To effectively generalize our model across different types of fire and environment, we must extend our studies to include a wider diversity of scenarios. (3) Integration of Investigator Behaviors: Incorporating data on fire investigator behaviors, potentially through eye-tracking technology, could provide new insights into the cognitive processes and decision-making patterns that influence the classification and origin determination tasks. This integration could further refine our tool, aligning it with practical investigative methods.

## Acknowledgment

This research was supported by the National Institute of Justice through the award #15PNIJ-22-GG-04442-RESS. We would like to express our sincere gratitude to Adam N. Friedman of the Bureau of Alcohol, Tobacco, Firearms and Explosives (ATF) for his invaluable assistance in facilitating the data collection process. His support was instrumental in ensuring the successful completion of this project.